\documentclass[lettersize,journal]{IEEEtran}

\usepackage{amsmath,amsfonts}
\usepackage{algorithmic}
\usepackage{algorithm}
\usepackage{array}
\usepackage[caption=false, font=footnotesize]{subfig}
\usepackage{textcomp}
\usepackage{stfloats}
\usepackage{url}
\usepackage{verbatim}
\usepackage{cite}
\usepackage{graphicx}
\usepackage{epstopdf}
\usepackage[utf8]{inputenc}
\usepackage[T1]{fontenc}
\usepackage{booktabs} 
\usepackage{diagbox}  
\usepackage{multirow} 
\usepackage{threeparttable} 
\usepackage[dvipsnames]{xcolor} 
\usepackage{color} 
\usepackage{soul}
\usepackage{orcidlink}
\hypersetup{hidelinks}
\soulregister{\cite}7 
\soulregister{\citep}7 
\soulregister{\citet}7 
\soulregister{\ref}7 
\soulregister{\pageref}7 
\usepackage{colortbl}  % cellcolor
\usepackage{hyperref}  % hyperlinks
\definecolor{codegreen}{rgb}{0,0.5,0}
\definecolor{codeblue}{rgb}{0,0,0.9}
\definecolor{codeblues}{rgb}{0,0,0.4}
\definecolor{codegray2}{rgb}{0.4,0.4,0.4}
\definecolor{codegray}{rgb}{0.9,0.9,0.9}

\newcounter{ALC@tempcntr}% Temporary counter for storage
\newcommand{\LCOMMENT}[1]{%
    \setcounter{ALC@tempcntr}{\arabic{ALC@rem}}% Store old counter
    \setcounter{ALC@rem}{1}% To avoid printing line number
    \item //#1 % Display comment + does not increment list item counter
    \setcounter{ALC@rem}{\arabic{ALC@tempcntr}}% Restore old counter
}%

\begin{document}
\title{Enhance Sample Efficiency and Robustness of End-to-end Urban Autonomous Driving via Semantic Masked World Model}

\author{

Zeyu Gao$^{\orcidlink{0000-0001-6771-5885}}$, Yao Mu$^{\orcidlink{0000-0002-0321-021X}}$, Chen Chen$^{\orcidlink{0000-0003-4310-8428}}$, Jingliang Duan$^{\orcidlink{0000-0002-3697-1576}}$, Ping Luo$^{\orcidlink{0000-0002-6685-7950}}$,~\IEEEmembership{Member,~IEEE}, Yanfeng Lu$^{\orcidlink{0000-0003-4601-4996}}$,~\IEEEmembership{Member,~IEEE}, and Shengbo Eben Li$^{\orcidlink{0000-0003-4923-3633}}$,~\IEEEmembership{Senior Member,~IEEE}

\thanks{

This work is supported by National Key Research and
Development Plan of China under Grant 2020AAA0105900, Tsinghua University Initiative Scientific Research Program, NSF China under Grants U20A20334, 52072213, 91948303, Beijing NSF
under Grant L211023. \textit{(Zeyu Gao and Yao Mu are co-first authors.)} \textit{(Corresponding author: Shengbo Eben Li and Yanfeng Lu.)}
\quad

Zeyu Gao and Yanfeng Lu are with the State Key Laboratory of Multimodal Artificial Intelligence Systems, Institute of Automation, Chinese Academy of Science, Beijing 100190, China, and also with the School of Artificial Intelligence, University of Chinese Academy of Sciences, Beijing 100049, China. (e-mail: gaozeyu2023@ia.ac.cn, yanfeng.lv@ia.ac.cn)
\quad 

Yao Mu and Ping Luo are with the Department of Computer Science, The University of Hong Kong, Hong Kong 999077, China. (e-mail: muyao@connect.hku.hk)
\quad 

Chen Chen and Shengbo Eben Li are with the State Key Laboratory of Automotive Safety and Energy, School of Vehicle and Mobility, Tsinghua University, Beijing 100084, China. (e-mail: lishbo@tsinghua.edu.cn)
\quad 

Jingliang Duan is with the School of Mechanical Engineering, University of Science and Technology Beijing, Beijing 100083, China. (e-mail: duanjl@ustb.edu.cn)
\quad 
}
}

\maketitle

%===============================================================================

\begin{abstract}
End-to-end autonomous driving provides a feasible way to automatically maximize overall driving system performance by directly mapping the raw pixels from a front-facing camera to control signals. Recent advanced methods construct a latent world model to map the high dimensional observations into compact latent space. However, the latent states embedded by the world model proposed in previous works may contain a large amount of task-irrelevant information, resulting in low sampling efficiency and poor robustness to input perturbations. Meanwhile, the training data distribution is usually unbalanced, and the learned policy is challenging to cope with the corner cases during the driving process. To solve the above challenges, we present a \textbf{SEM}antic \textbf{M}asked recurrent world model (\textbf{SEM2}), which introduces a semantic filter to extract key driving-relevant features and make decisions via the filtered features, and is trained with a multi-source data sampler, which aggregates common data and multiple corner case data in a single batch, to balance the data distribution. Extensive experiments on CARLA show our method outperforms the state-of-the-art approaches in terms of sample efficiency and robustness to input permutations.
\end{abstract}

% Two or three meaningful keywords should be added here

%===============================================================================

\section{Introduction}
\label{sec:Introduction}

\IEEEPARstart{E}{nd-to-end} autonomous driving learns to map the raw sensory data directly to driving commands via deep neural networks. Compared to the traditional modularized-learning framework, which decomposes the driving task into lane marking detection~\cite{neven2018towards,hou2019learning}, path planning~\cite{okamoto2019optimal}, decision making\cite{hubmann2017decision} and control~\cite{chen2019autonomous,duan2023relaxed}, the end-to-end method aims to immediately maximize overall driving system performance. The advantages of end-to-end autonomous driving are two folds. Firstly, the internal components self-optimize to maximize the use of sample information toward the best system performance instead of optimizing human-selected intermediate criteria, e.g., lane detection. Additionally, smaller networks are possible since the system learns to solve the problem with a minimal number of processing components. There are two mainstreams of the end-to-end method, imitation learning~\cite{bojarski2016end,pan2017agile,codevilla2018end,xiao2020multimodal} and reinforcement learning~\cite{wang2018deep,chen2019model,wu2022uncertainty,chen2021interpretable}. Imitation learning learns the driving policy from large amounts of human driving data, and the policy performances are upper limited by drivers' capacities and scenario diversity. Contrarily, the reinforcement learning method enables the agent to learn the optimal policy by fully interacting with the environments and hence reduces the need for expert data, and is increasingly applied to end-to-end autonomous driving with extraordinary performances. 

However, current end-to-end methods have been subject to criticism due to two primary issues: 1) Learning driving policy directly from high-dimensional sensor input is challenging, as known as dimensional disasters. 2) The majority of existing approaches concentrate on straightforward driving tasks, such as lane-keeping, but exhibit inadequate performance in complex urban scenarios. The variety of urban scenes and street landscapes greatly increases the complexity of the samples, making it challenging to learn an efficient and robust end-to-end driving policy. Recent advanced end-to-end autonomous driving methods build latent world models to abstract high-dimensional observations into compact latent states, enabling the self-driving vehicle to predict forward and learn from low-dimensional states. Chen. et al. introduce an end-to-end self-driving framework with a stochastic sequential latent world model to reduce the sample complexity of reinforcement learning~\cite{chen2021interpretable}. LVM~\cite{zhang2021steadily} further introduces Dreamer~\cite{hafner2020dream,hafner2021mastering}, a novel recurrent world model with a deterministic path and a stochastic path concurrently, into the autonomous driving framework to improve prediction accuracy and the stability of the driving policy learning process. However, these world models still contain significant amounts of driving-irrelevant information, such as clouds, rain, and surrounding buildings, leading to low sampling efficiency and limiting robustness to input perturbations. 
Additionally, the data distribution for the training world model is usually unbalanced, i.e., straight-line driving data appears more while turning, and near-collision data is comparatively less, making the agent face difficulties in handling corner cases and may even cause modal collapse. Can we train a world model that extracts driving-relevant features from a balanced sample distribution, and construct an efficient and robust end-to-end autonomous driving framework, anticipating its outstanding performances in both common scenes and corner cases?

To overcome the aforementioned challenges, this paper proposes a \textbf{SEM}antic \textbf{M}asked recurrent world model (\textbf{SEM2}) to enhance the sample efficiency and robustness of the autonomous driving framework. The SEM2 allows the agent to concentrate solely on task-related information via a latent feature filter learned by reconstructing the semantic mask. The semantic mask provides driving-relevant information consisting of the road map, the target path, and surrounding objects in a bird-view form, formulating a semantic masked world model to obtain the transition dynamics of the driving-relevant latent state. Then, the agent learns the optimal policy by taking the filtered features as input, and the learned optimal policy can generate highly driving-tasks-correlated actions, hence it would be more robust to input permutations compared with previous works. To tackle the uneven data distribution issue, we collect the common data and corner case data separately and construct a multi-source data sampler to aggregate different scenes in a mini-batch for the training of the semantic masked world model.
Key contributions of our work are summarized as follows:
	
\begin{itemize} 
	\item{We propose a \textbf{SEM}antic \textbf{M}asked recurrent world model (\textbf{SEM2}) that learns the transition dynamics of driving-relevant states through a semantic filter to reduce the interference of irrelevant information in sensor inputs, thus improving sampling efficiency and robustness of learned driving policy.}
	\end{itemize}

	\begin{itemize}
	\item{A multi-source data sampler is proposed to balance data distribution and prevent modal collapse in corner cases, which contributes diverse scene data in training the semantic masked world model by using both common driving situations and multiple corner cases in urban scenes.}
	\end{itemize}
	
	\begin{itemize} 
	\item{Extensive experiments conducted on the CARLA benchmark show our method surpasses the previous works of end-to-end autonomous driving with deep reinforcement learning in terms of sample efficiency and robustness to input permutation.}
	\end{itemize}
	
%===============================================================================

\section{Related Works}
\label{sec:Related Works}
\subsection{End-to-end Autonomous Driving}
Prior research on end-to-end autonomous driving can be categorized into two primary branches: imitation learning (IL) and reinforcement learning (RL). IL learns a driving policy by mimicking the behavior of expert drivers~\cite{bojarski2016end,pan2017agile,codevilla2018end,xiao2020multimodal}. Agents can usually learn good driving strategies by imitating human driving behavior. However, this is similar to the modularized framework in essence and does not break through the limitations of human experience. RL collects data through the interaction between the agent and environment and learns from data. In recent years, RL has developed rapidly and gets a series of achievements~\cite{mnih2015human,levine2016end,silver2016mastering,silver2017mastering,haarnoja2018soft,duan2021distributional}. In the previous work, the model-free method was used to complete the end-to-end autonomous driving task (e.g. DQN on Gazebo~\cite{wolf2017learning}, DDPG on TORCS~\cite{wang2018deep}, SAC on CARLA~\cite{chen2019model}, DQN on CARLA~\cite{yurtsever2020integrating}). For the sake of sample efficiency, most recent studies use model-based methods (e.g. UA-MBRL~\cite{wu2022uncertainty}, LVM~\cite{zhang2021steadily}, GCBF~\cite{ma2021model}). In addition, interpretability is also the focus of the work. Some researches are devoted to improving the interpretability of agents (e.g. visual explanations~\cite{kim2017interpretable}, semantic birdeye mask~\cite{chen2021interpretable}, interpretable learning system~\cite{guan2021integrated}). Despite previous efforts, there remains scope for improvement in terms of optimizing the utilization of high-dimensional semantic information and balancing data distribution.

\subsection{Latent World Model}
In dealing with high-dimensional inputs, the latent world model provides a flexible way to represent the key information of observations. The world model~\cite{ha2018recurrent} is learned through two stages: representation learning and latent dynamics learning. Transforming the high-dimensional inputs into compact state representations, this model shows its superiority in doing numerous predictions in a single batch, without the need for image generation. PlaNet~\cite{hafner2019learning} coordinated the two stages and proposed recurrent stochastic state model (RSSM), which enables fast online planning in latent space with both deterministic and stochastic components. PlaNet-Bayes~\cite{okada2020planet} represented an advancement of the PlaNet model by incorporating Bayesian inference to account for the presence of two forms of uncertainty in the latent models and actions. Dreamer~\cite{hafner2020dream} can use the RSSM model to imagine the long-term future. DreamerV2's~\cite{hafner2021mastering} latent variable uses vectors of multiple classification variables and optimizes them using straight-through gradients. Stochastic latent SAC (SLAC)~\cite{lee2020stochastic} offers a principled approach to merge stochastic sequential models and RL into a unified method. Bridging reality and dream (BIRD)~\cite{zhu2020bridging} aims to improve policy generalization by maximizing mutual information between real and imaginary trajectories. The advancement of the latent model presents novel opportunities for the realization of end-to-end autonomous driving systems.

%===============================================================================

\section{Preliminary}
   
\subsection{Reinforcement Learning for Autonomous Driving}
In the context of driving environments, a reinforcement learning agent strives to learn an optimal policy that maximizes cumulative rewards by exploring Markov decision processes (MDP). In general, the time step is denoted by $t$, and we introduce the state $s_t \in \mathcal{S}$, action $a_t \in \mathcal{A}$, reward function $r(s_t, a_t)$, policy $\pi_{\theta}(s)$, and transition probability $p(s_{t+1}|s_t, a_t)$ to characterize the interaction process with the environment in reinforcement learning. The reward function is usually composed of driving efficiency, driving compliance, safety, and energy efficiency. The goal of the agent is to find policy parameters $\theta$ that maximize the long-horizon summed rewards represented by a value function $v_\varphi(s_t)\doteq\mathbb{E}\left(\sum\limits_{i=t}^{t+H}\gamma^{i-t}r_{i} \right)$ parameterized with $\varphi$. In advanced RL-based autonomous driving framework, the agent constructs a world model $p_\phi$, with $\phi$ as its parameterization, to represent the environmental dynamics $p$ and the reward function $r$, and then performs planning or policy optimization based on the imagination with the learned world model to achieve efficient and safe driving performance.

\begin{figure}[t]
		\centering
		\includegraphics[width=0.45\textwidth]{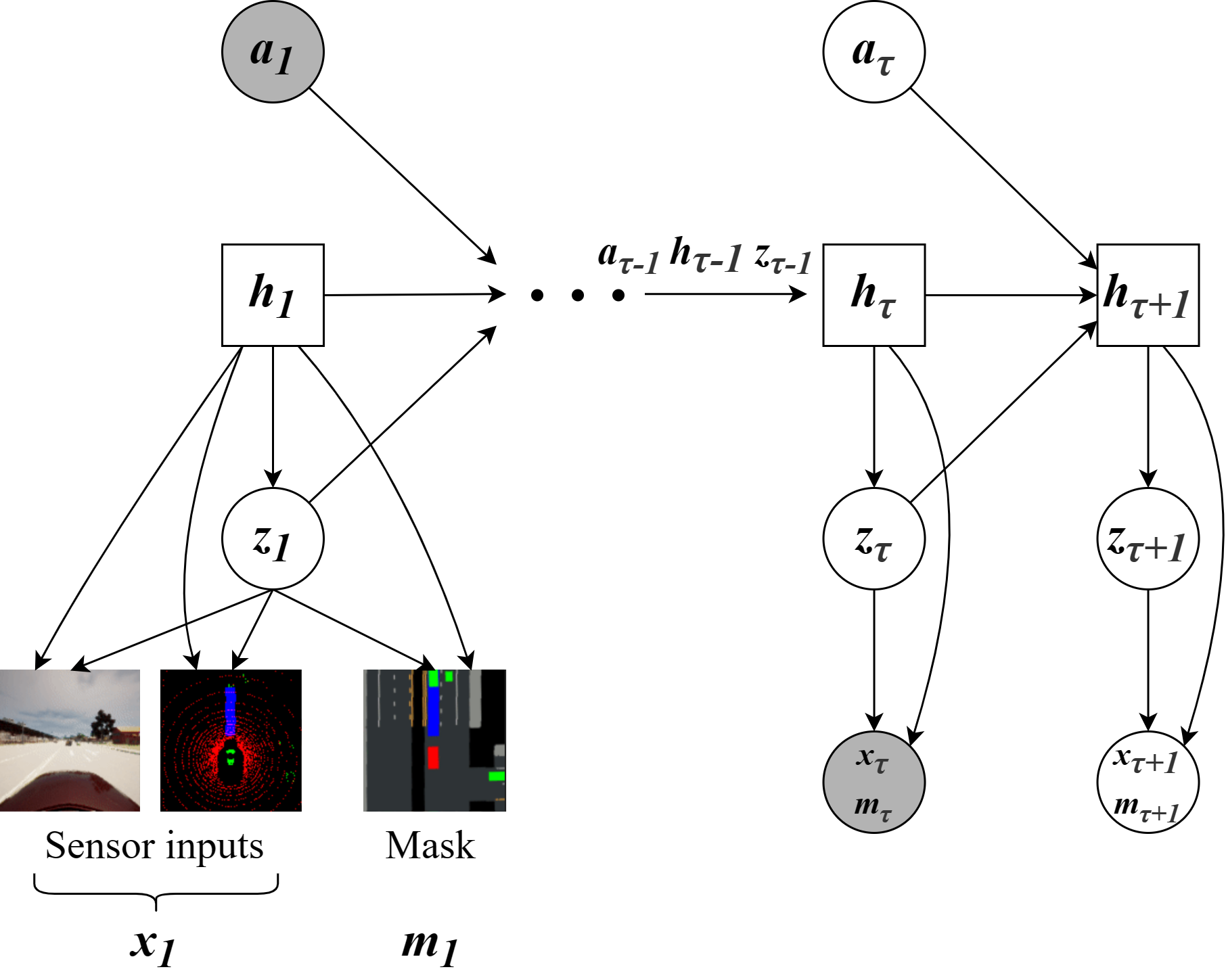}
		\caption{\label{rssm-model} Structure of the recurrent state space model (RSSM). The latent state $s_{i}$ in RSSM is composed of a deterministic variable $h_{i}$ and a stochastic variable $z_{i}$. The generative process is represented by solid lines, while the inference model is represented by dashed lines. The latent dynamic contains a deterministic path and a stochastic path, which learns historical information efficiently while accurately modeling the stochastic characteristic of the dynamics. }
\vspace {-0.5cm}
\end{figure}
    
\subsection{World Model}
In this work, we consider sequences $\{o_t,a_t,r_t\}_{t=1}^T$ where $t$ denotes the discrete time step, $o_t$ is the sensor observation, $a_t$ is the continuous action vector, and $r_t$ is the scalar reward. A typical recurrent latent state-space model (RSSM) is shown in Fig.~\ref{rssm-model} and resembles the structure of a partially observable Markov decision process (POMDP). In RSSM, the latent state $s_t$ is partitioned into a deterministic component $h_t$ and a stochastic component $z_t$, and predicts the future states $s=\{h_t,z_t\}_{t=1}^T$ with both the deterministic path and stochastic path. The parameters of the RSSM are optimized by maximizing the variational bound using Jensen's inequality:
% \begin{equation}
% \begin{aligned}
% &\ln p\left(o_{1: T} \mid a_{1: T}\right) = \ln \int \prod_{t} p\left(s_{t} \mid s_{t-1}, a_{t-1}\right) p\left(o_{t} \mid s_{t}\right) \mathrm{d} s_{1: T} \\
% &\left.\geq \sum_{t=1}^{T}\left(\mathrm{E}_{q\left(s_{t} \mid o_{\leq t}, a_{<t}\right) } \ln p\left(o_{t} \mid s_{t}\right)\right. \right. \rightarrow \\
% &\quad-\mathrm{\underset{{q\left(s_{t-1} \mid o_{\leq t-1}, a_{<t-1}\right)}\hfill}{E\big[\operatorname{KL}\left[q\left(s_{t} \mid o_{\leq t}, a_{<t}\right) \| p\left(s_{t} \mid s_{t-1}, a_{t-1}\right)\right]\big]}}\Big)
% \end{aligned}
% \end{equation}
\begin{small}
\begin{equation}
\begin{aligned}
&\ln p\left(o_{1: T} \mid a_{1: T}\right) = \ln \int \prod_{t} p\left(s_{t} \mid s_{t-1}, a_{t-1}\right) p\left(o_{t} \mid s_{t}\right) \mathrm{d} s_{1: T} \\
&\left.\geq \sum_{t=1}^{T}\left(\mathrm{E}_{q\left(s_{t} \mid o_{\leq t}, a_{<t}\right) } \ln p\left(o_{t} \mid s_{t}\right)\right. \right. - \\
&\quad\mathop{\mathrm{E}}_{\scriptsize q\left(s_{t-1} \mid o_{\leq t-1}, a_{<t-1}\right)}\big[\operatorname{KL}\left[q\left(s_{t} \mid o_{\leq t}, a_{<t}\right) \| p\left(s_{t} \mid s_{t-1}, a_{t-1}\right)\right]\big]\Big).
\end{aligned}
\end{equation}
\end{small}

Given the learned world model, the efficient utilization of neural network latent dynamics enables the acquisition of long-term behaviors in a compact latent space. To this end, multi-step returns gradients are propagated through the world model predictions of actions, states, rewards, and values using reparameterization. We use the notation $i$ to represent the time index for imagined quantities. Imagined trajectories start at initial state $s_t$ and follow predictions made by the world model $s_i \sim p(s_i|s_{i-1},a_{i-1})$, reward model $r_i \sim p(r_i|s_i)$, and a policy $a_i \sim \pi(a_i|s_i)$. The objective is to maximize expected imagined rewards $\mathrm{E}_{\pi}\left(\sum\limits_{i=t}^{\infty} \gamma^{i-t} r_{i}\right)$ with respect to the policy.

\begin{figure*}[t]
% \vspace{-15pt}
		\centering
		\includegraphics[width=0.85\textwidth]{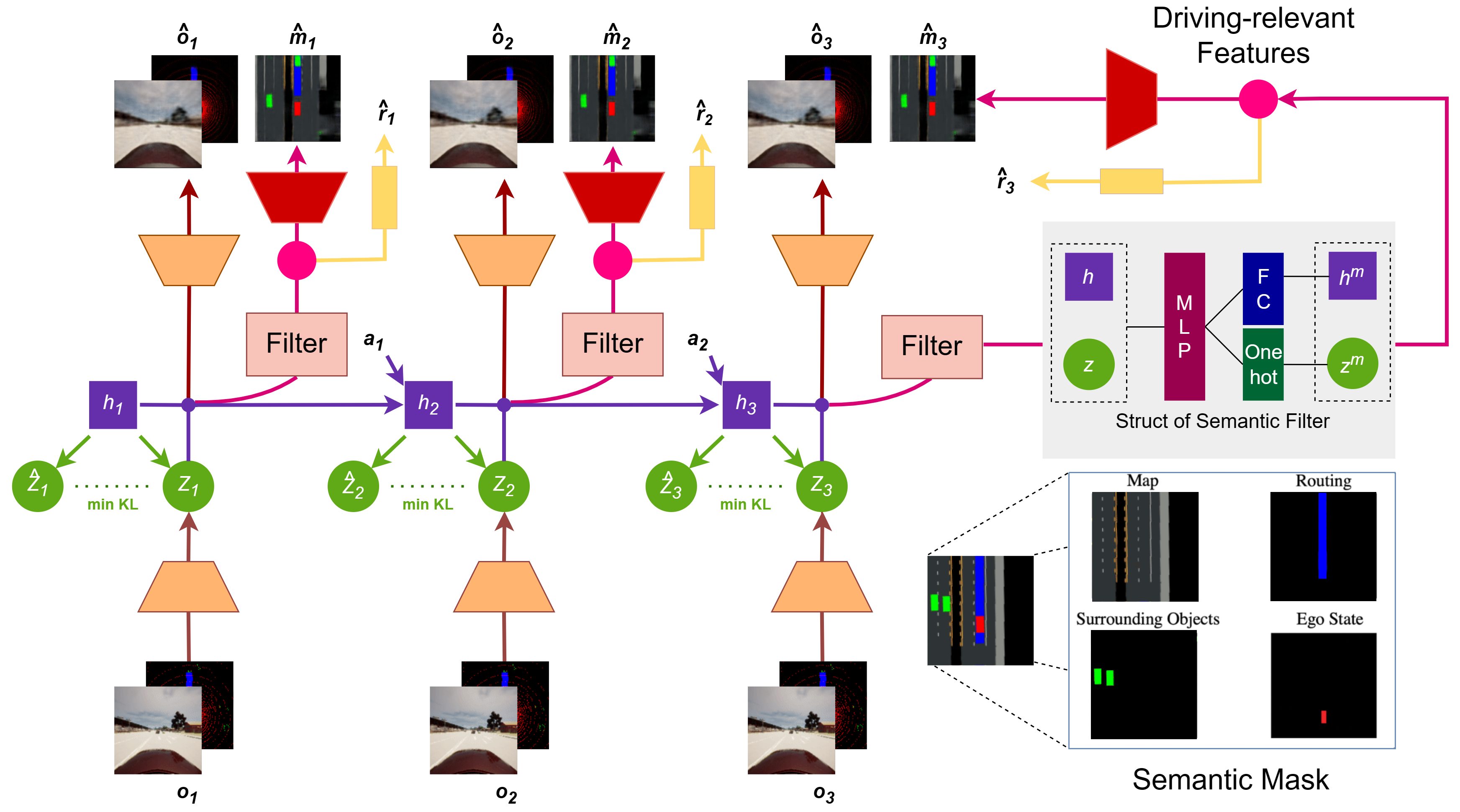}
    \vspace{-5pt}
		\caption{\label{3} The overall structure of SEM2. SEM2 takes the observation $o_{t}$ from the camera and lidar as input and then encodes it into the latent state which contains deterministic variable $h_{t}$ and stochastic variable $z_{t}$. The original features are used to reconstruct the observation. The latent semantic filter extracts the driving-relevant features from the original features and reconstructs the semantic mask $\hat{m}_{t}$ and predicts the reward $\hat{r}_{t}$.}
  \vspace{-10pt}
\end{figure*}

\section{Semantic Masked World Model}
The world model condenses the agent's experience into a predictive model, making it possible to predict multiple trajectories simultaneously over a long period without having to generate images. To enhance the sample efficiency and robustness, the world model learns the transition dynamics of the latent states that only pay attention to the driving-relevant features via semantic filter, thus reducing the interference of driving-irrelevant information of sensor inputs in urban scenes. 

\subsection{Model Structure}
As shown in Fig. \ref{3}, SEM2 consists of i) a recurrent model to extract useful information from historical information, which is implanted as a typical recurrent neural network gate recurrent unit (GRU)\cite{cho2014learning}, ii) a representation model to encode the observation from sensor input into the latent space, iii) a transition predictor to predict the state transition, iv) a semantic filter to extract driving-relevant features, v) a mask predictor to reconstruct the semantic birdeye mask, vi) an observation predictor to reconstruct the observation and vii) a reward predictor to predict the reward given by the environment. The detailed structure of model components can be represented as:
	\begin{equation}
        \begin{array}{lll}
        \text { Recurrent model: } &  &h_{t} =f_{\phi}\left(h_{t-1}, z_{t-1}, a_{t-1}\right) \\
        \text { Representation model: } & & z_{t} \sim q_{\phi}\left(z_{t} \mid h_{t}, o_{t}\right) \\
        \text { Transition predictor: } & & \hat{z}_{t} \sim p_{\phi}\left(\hat{z}_{t} \mid h_{t}\right) \\
        \text{ Semantic filter: } & & s^m_t \sim S_\phi(s^m_t|h_t,z_t)\\
        \text{ Mask predictor: } & & \hat{m}_t \sim p_\phi(\hat{m}_t|s^m_t)\\
        \text { Observation predictor: } & & \hat{o}_{t} \sim p_{\phi}\left(\hat{o}_{t} \mid h_{t}, z_{t}\right) \\
        \text { Reward predictor: } & & \hat{r}_{t} \sim p_{\phi}\left(\hat{r}_{t} \mid s^{m}_{t}\right). \\
        \end{array}
    \end{equation}
    
All components are implemented as neural networks, with $\phi$ denoting their amalgamated parameter vectors. The representation model is realized by convolutional neural networks (CNNs)~\cite{lecun1989backpropagation} followed by a multi-layer perceptron (MLP) that takes the image embedding and deterministic recurrent state as input. The observation and mask predictor are transposed CNNs. The transition and reward predictor are MLPs. The transition predictor only predicts the next latent state based on the historical information, current state, and action, without using the next observation. In this way, we can predict future behavior without the need to observe or generate observations. 

\subsection{Semantic Filter and Mask}
The autonomous driving system takes observations from the front camera and lidar as inputs and encodes the high-dimensional inputs into the latent space. Since there is often a large amount of driving irrelevant information in the camera images, such as weather, sky, and tall buildings, the raw feature encoding is susceptible to interference from irrelevant information, which reduces the robustness of the learned policy. For example, when the vehicle encounters rainstorm conditions, the raw feature encoding may change and directly impact the action of the agent.

An effective strategy for solving the aforementioned problem is to establish a strong correlation between decisions and high-dimensional semantic masks. However, merely reconstructing the high-dimensional semantic mask is not sufficient to achieve this goal. It is necessary to include a component that separates the reconstruction process of observation and semantic mask. This separation enables the efficient use of the semantic mask and extracts task-relevant features, thereby improving the robustness and performance.

\begin{figure}[ht]
    \centering
    \includegraphics[scale=0.1]{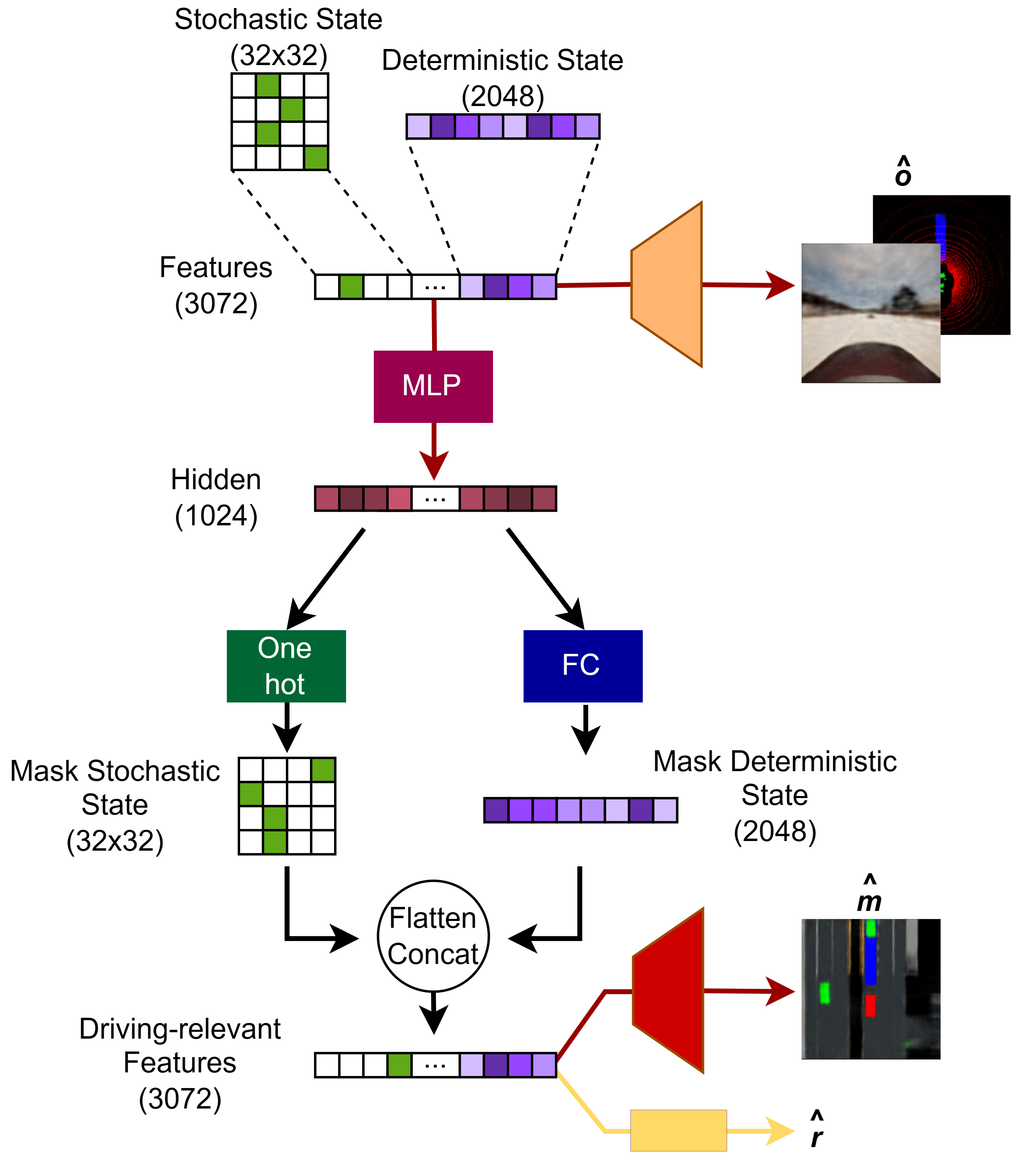}
    \caption{\label{semantic_filter} Working process of semantic mask filter.}
\end{figure}

As shown in the upper right of Fig. \ref{3}, we introduce semantic filter $\mathcal{S}_\phi(\cdot|\cdot)$ which is the key component to extract the driving-relevant features. The semantic filter, which is implanted as an MLP structure, takes the latent features inferred by the recurrent neural network as input and extracts the driving-relevant features as output to generate the semantic mask and all driving-relevant variables, such as reward and action. To ensure that the driving-relevant features are consistent with the original feature form, we use the fully connected (FC) head and the categorical head to separately output deterministic and stochastic states. The working process of the semantic mask filter is shown in Fig \ref{semantic_filter}. 
The semantic mask, as shown in the lower right of Fig. \ref{3}, contains comprehensive driving-relevant information in a bird-view that can be understood by humans, which includes the maps that represent road features, routing that represents the road a vehicle aims for, state of surrounding vehicles, and the ego state that represents the state of the ego vehicle. 

\subsection{Loss Function} 
The constituent elements of SEM2 are optimized jointly to maximize the variational lower bound, as suggested in the work of Hafner et al. \cite{hafner2021mastering}. The optimization aims to train the distributions generated by the transition predictor, the observation predictor, the mask predictor, and the reward predictor by maximizing the log-likelihood of their corresponding targets. The mask predictor reconstructed the semantic bird-view mask via the filtered features $s^{m}_{t}$. Thus the semantic filter $S$ is optimized with the part of mask log loss to minimize the error between the reconstructed mask and ground truth obtained directly from the simulator. The loss function of the SEM2 can be derived as:
\begin{equation}
\footnotesize
\label{World model loss}
\begin{aligned}
& \ln{p\left( o_{1:T},m_{1:T},r_{1:T} \right)}\geq \mathcal{L}(o_{1:T},m_{1:T},r_{1:T};\phi){\overset{.}{=}}\quad\quad\quad\quad\quad\quad\quad\quad\quad\quad\quad\quad\quad\quad\\
% \underset{{q_{\phi}(s_{1:T}|a_{1:T},o_{1:T})}}{\mathrm{E}}\left\lbrack {\sum\limits_{t = 1}^{T}\underset{\text{image~log~loss}}{\underbrace{- {\ln p_{\phi}}\left( o_{t} \middle| {h_{t},z_{t}} \right)}}}\underset{\text{mask~log~loss}}{\underbrace{- {\ln p_{\phi}}\left( m_{t} \middle| {S_\phi(h_{t},z_{t})}\right)}}
% \right.\\\left.
% \underset{\text{reward~log~loss}}{\underbrace{- {\ln p_{\phi}}\left( r_{t} \middle| {S_\phi(h_{t},z_{t})} \right)}}\underset{\text{KL~loss}}{\underbrace{+ {{~KL\left\lbrack q_{\phi}\left( z_{t} \middle| {h_{t},o_{t}} \right) \middle| \middle| p_{\phi}\left( z_{t} \middle| h_{t} \right) \right\rbrack}}}} \right\rbrack.
& \underset{\begin{matrix}
{{h_{1:T},z_{1:T} \sim q}_{\phi}{({h_{1:T},z_{1:T}|a_{1:T},o_{1:T}})}~} \\
{s_{1:T}^{m} \sim S_{\phi}{({s_{1:T}^{m}|h_{1:T},z_{1:T}})}}\end{matrix}}{E}\left\lbrack {{\sum\limits_{t = 1}^{T}~}\underset{observation~loss}{\underbrace{- {\ln{p_{\phi}\left( o_{t} \middle| h_{t},z_{t} \right)}}}}} \right. \\
& \left. {\underset{mask~loss}{\underbrace{- {\ln{p_{\phi}\left( m_{t} \middle| s_{t}^{m} \right)}}}}\underset{reward~loss}{\underbrace{- {\ln{p_{\phi}\left( r_{t} \middle| s_{t}^{m}\right)}}}}\underset{KL~loss}{\underbrace{+ KL\left\lbrack q_{\phi}\left( z_{t} \middle| h_{t},o_{t} \right) \middle| \middle| p_{\phi}\left( z_{t} \middle| h_{t} \right) \right\rbrack}}} \right\rbrack
\end{aligned}
\end{equation}

The architecture of SEM2 can be interpreted as a sequential variational autoencoder (VAE), where the representation model acts as an approximate posterior and the transition predictor acts as the temporal prior. The Kullback-Leibler (KL) loss \cite{joyce2011kullback} serves a dual purpose: it guides the prior towards the representations, and it regulates the representations towards the prior. The driving-relevant features $s_{t}^{m}=\{h^m_t,z^m_t\}$ extracted by latent semantic filters are used as the final representation, which aggregates historical information and current observations while serving as inputs to the policy network and value network.

\subsection{Multi-source Sampler} 
During the data collection phase, self-driving vehicles primarily accumulate data while navigating straight roads with low traffic density. However, data collected in scenarios involving curves and densely populated traffic environments is limited. This leads to suboptimal performance in reconstructing the mask by the world model and the policy tends to perform poorly in these corner cases.

\begin{figure}[ht]
    \centering
    \includegraphics[scale=0.13]{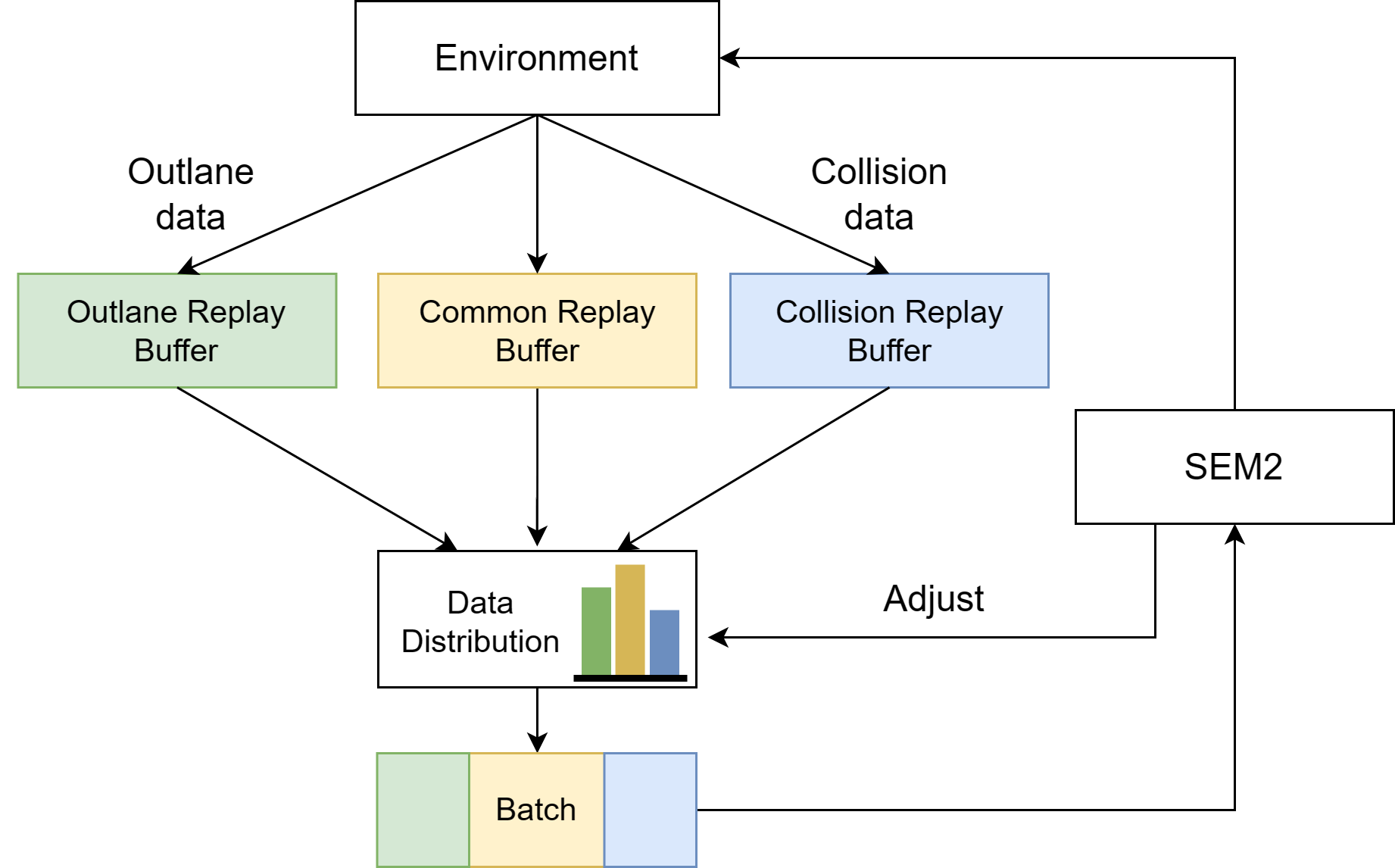}
    \caption{\label{5} The structure of the multi-source sampler for the training of SEM2. In addition to the common replay buffer, there are two corner case replay buffers that save the data in outlane cases and collision cases independently. In every iteration of the training process, we sample mini-batch from the three replay buffers, in turn, to contribute diverse data to support the SEM2 updating. The data distribution will be adjust after the evaluation of SEM2.}
\end{figure}

To address this gap, we propose a multi-source training method. This method utilizes a multi-source sampler which is illustrated in Fig. \ref{5} to mix batches. The multi-source sampler separately collects the abnormal ending episodes in the process of interaction with the environment as the boundary condition data set. We divide the replay buffer into three categories: common replay buffer, outlane replay buffer, and collision replay buffer. Each time the system is trained, the corner case data set is used by the multi-source sampler to train the world model. The agent with multi-source training can better cope with the corner case, obtain higher average returns and improve driving performance.

An additional challenge focuses on how to balance the distribution of both common cases and corner cases. To address this challenge, we introduce an adaptive method to adjust the data distribution. We record checkpoints at every evaluation step (20,000) of the agent interacting with the environment to evaluate the current driving performance of the agent with episodes (30) and record the success rate, collision rate and outlane rate. In the initial stages of training, the agent necessitates a higher proportion of common data to learn the model and policy. As the training regimen progresses, the agent's success rate incrementally ascends, demanding an augmented allocation of corner data to effectively learn to address these corner scenarios. The ratio of collision data to outlane data within the corner data corresponds equivalently to the ratio of collision rate to outlane rate. The proportion of corner data should not be too large to prevent overfitting and an overly conservative policy. We limit the proportion of corner data by a hyperparameter called max corner data probability $P_{\text{maxcor}}$. Alg. \ref{adaptive adjustment} describes in detail the process of adaptive adjustment of data distribution. By doing so, the distribution of training data is balanced, which enhances the adaptability of the world model and the policy in handling corner cases.

In the training process of SEM2, the batches of $B = 16$ sequences are sampled from the adaptive data distribution, each having a fixed length of $L = 16$. 
	
\begin{algorithm}
    \caption{Adaptive adjustment of data distribution}
    \label{adaptive adjustment}
    \begin{algorithmic}[1]
        \STATE Initialize common data sampling probability $P_{com}=1$, collision data sampling probability $P_{col}=0$, outlane data sampling probability $P_{out}=0$, and max corner data probability $P_{cormax}$.
        \WHILE{not converged}
            \STATE use the multi-source data sampler to train SEM2 and collect multi-source data.
            \IF {time step $t$ is divisible by evaluation step $t_{eval}$}
                \STATE run 30 episodes and record success rate $r_{suc}$, collision rate $r_{col}$, outlane rate $r_{out}$.
                \STATE update multi-source data sampler's parameters $P_{cor} = r_{suc} \times P_{cormax}$, $P_{com}=1-P_{cor}$, $P_{col} = \frac{r_{col}}{r_{col}+r_{out}}  P_{cor}$, $P_{out} = \frac{r_{out}}{r_{col}+r_{out}} P_{cor}$.
            \ENDIF
        \ENDWHILE    
    \end{algorithmic}
\end{algorithm}

%===============================================================================
\section{Behavior Learning }
\label{sec: behavior learning }
\subsection{Latent imagination}
\begin{figure}
    \centering
    \includegraphics[width=0.498\textwidth]{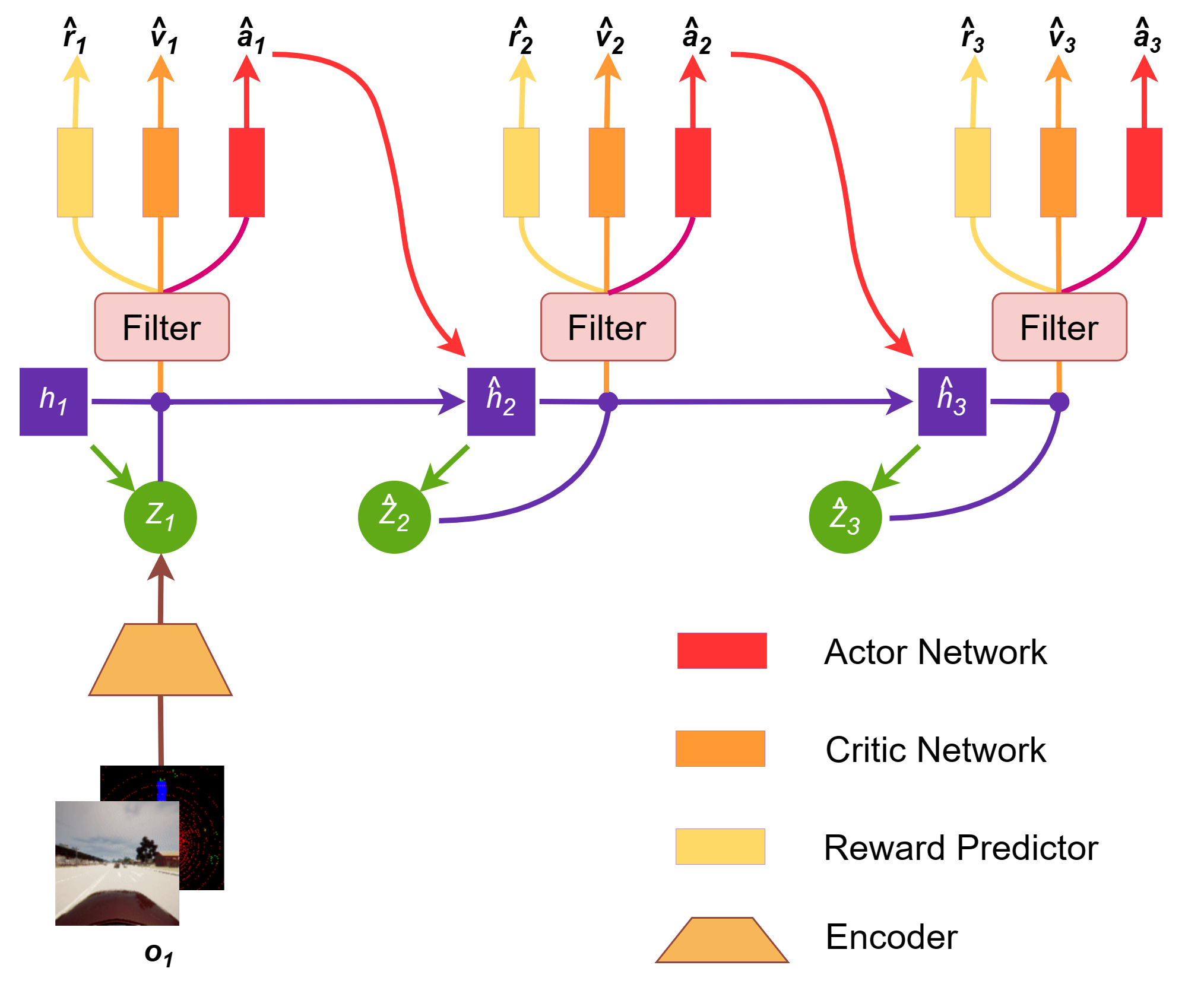}
    \caption{\label{4} The semantic masked world model is utilized to acquire knowledge of the policy from sequences imagined in the condensed latent space. These sequences commence from posterior states computed during the model training and advance by generating actions sampled from the actor that possesses filtered driving-relevant features. The critic network estimates the expected rewards for each state by applying temporal difference learning to the imagined rewards. The actor is trained to maximize the expected rewards via the straight-through gradients of the learned world model. }
\end{figure}

We aim to learn smooth and safe driving policy through long-term imaginary trajectories unrolled by the learned world model SEM2 with high sample efficiency. For this, multi-step returns gradients are propagated through the world model predictions of actions, states, rewards, and values using reparameterization with the help of SEM2.
As shown in Fig. \ref{4}, we learn long-term behavior with SEM2 by the imagery process, which predicts the latent states of the next $I$ steps from an initial state. The initial state of the process is obtained from the input images, and the subsequent hidden variables $\hat{h},\hat{z}$, driving-relevant features $\hat{s}^m$, actions $\hat{a}$ and rewards $\hat{r}$ are obtained from the world model predictions. 

\subsection{Actor-Critic framework}
The optimal policy is learned within the context of an actor-critic framework utilizing the SEM2 model, which is comprised of two components: an actor responsible for selecting actions that maximize the anticipated planning reward, and a critic that predicts future rewards.

We use a stochastic actor that selects actions and a deterministic critic to learn long-term behaviors in the imagination process. The actor learns to generate actions according to the filtered latent state $s^{m}$ to maximize the state value predicted by the critic, which learns to estimate the actor's cumulative rewards. The actor $\pi_\theta$ and critic ${v_\varphi}$ are represented respectively as:
\begin{equation}
\begin{split}
    \text{Actor: }\quad & \hat{a}_t \sim \pi_\theta(\hat{a}_t|s^m_t),\\
   \text{Critic: }\quad & v_\varphi(\hat{s}^m_t) {\overset{.}{=}} \mathrm{E}(\sum\limits_{i = t}^{t+I}\gamma^{i-t}\hat{r}_i).
\end{split}
\end{equation}

Since the features fed directly to the actor will directly determine the quality of the generated action, in SEM2 we utilize the driving-relevant feature $s^{m}_{t}=\{h^m_t,z^m_t\}$ generated by the latent semantic filter as the input of the actor, which is highly correlated with the semantic mask and greatly reduces the interference of irrelevant information. Through this sequential process, the agent learns to update the parameters of actor and critic networks by the gradient descent method without changing the parameter of the SEM2 model. As the world model is held constant during behavior learning, gradients concerning the actor and critic do not influence its representations, allowing us to simulate a large number of latent trajectories efficiently on a single GPU.

The objective of the critic is to estimate the state value function, which is defined as the expected discounted sum of future rewards that the agent will receive. To this end, we use the TD-learning method, where the critic is trained to predict a value target, which consists of the intermediate rewards and the output of the critic in the later latent states. To take full advantage of the world model's ability to make multi-step predictions and to regularize the variance of the estimates, we use the TD-$\lambda$~\cite{schulman2015high} objective to learn the value function, which is recursively defined as follows:
\begin{equation}
\label{Critic loss}
\begin{split}
 &\mathcal{L(\varphi)} \doteq \mathrm{E}_{p_{\phi}, \pi_{\theta}}\left[\sum_{t=1}^{I-1} \frac{1}{2}\left(v_{\varphi}\left(\hat{s}^m_{t}\right)-\left(V_{t}^{\lambda}\right)\right)^{2}\right],\\
&V_{t}^{\lambda} \doteq \hat{r}_{t}+\hat{\gamma}_{t} \begin{cases}(1-\lambda) v_{\varphi}\left(\hat{s}^m_{t+1}\right)+\lambda V_{t+1}^{\lambda} & \text { if } t<I, \\ v_{\varphi}\left(\hat{s}^m_{I}\right) & \text { if } t=I.\end{cases}
\end{split}
\end{equation}

The TD-$\lambda$ target is computed as a weighted average of n-step returns for different horizons. In practice, we set $\lambda = 0.95$. Notably, the gradient of the value target $V_{t}^{\lambda}$ is stopped to restrict the critic loss $\mathcal{L(\varphi)}$ only updates the critic.

The actor aims to maximize the TD-$\lambda$ return predicted by the critic while regularizing the entropy of the actor to encourage exploration: 
\begin{equation}
\label{Actor loss}
\mathcal{L}(\theta) \doteq \mathrm{E}_{p_{\phi}, \pi_{\theta}}\left[\sum_{t=1}^{I-1}(\underbrace{V_{t}^{\lambda}}_{\text{dynamics}}-\underbrace{\eta \mathrm{H}\left[a_{t} \mid \hat{s}^m_{t}\right]}_{\text{entropy~regularizer}})\right].
\end{equation}

The actor and critic are implemented as MLPs with ELU~\cite{clevert2015fast}. They are trained using the same imagined trajectories but optimize separate loss functions.

With this, the exposition of the model learning and behavior learning processes concludes. We briefly introduce the SEM2 algorithm in Alg. \ref{SEM2 algorithm}. 

\begin{algorithm}
    \caption{SEM2}
    \label{SEM2 algorithm}
    \begin{algorithmic}[1]
        \STATE Initialize multi-source dataset $\{D_{com},D_{col},D_{out}\}$ by random policy.\\
        \STATE Initialize networks parameters $\phi$, $\varphi$, $\theta$.
        \WHILE{not converged}
            \FOR{update step $u$ = 1 \TO $U$}
                \LCOMMENT{ Learning the semantic masked world model }
                \STATE Sample $B$ data sequences by multi-source data sampler $\left\{ (a_{t-1},o_t,m_t,r_t)\right\}_{t=k}^{k+I}\sim \{D_{com},D_{col},D_{out}\}$. 
                \STATE Compute states ${h_t}=f_{\phi}\left(h_{t-1}, z_{t-1}, a_{t-1}\right)$ and ${z_t}\sim p_{\phi}\left(\hat{z}_{t} \mid h_{t}\right)$.
                \STATE Reconstruct observations $\hat{o}_{t} \sim p_{\phi}\left(\hat{o}_{t} \mid h_{t}, z_{t}\right)$.
                \STATE Compute driving-relevant features by semantic filter ${s^m_t \sim S_\phi(s^m_t|h_t,z_t)}$.
                \STATE Reconstruct semantic mask $\hat{m}_t \sim p_\phi(\hat{m}_t|s^m_t)$, ego vehicle state vector $\hat{e}_{t} \sim p_{\phi}\left(\hat{e}_{t} \mid s^{m}_{t}\right)$ and reward $\hat{r}_{t} \sim p_{\phi}\left(\hat{r}_{t} \mid s^{m}_{t}\right)$ to train semantic filter.
                \LCOMMENT{ Behavior learning}
                \STATE Imagine $\{\hat{s}_i = \{h_t,z_t\}\}_{i=t}^{t+I}$ from each state $s_t$, extract $\{\hat{s}^m_i\sim S_\phi(s^m_i|s_i)\}_{i=t}^{t+I}$, and predict trajectories of action $\{\hat{a}_i \sim \pi_\theta(\hat{a}_i|s^m_i)\}_{i=t}^{t+I}$,\quad value $\{v_\varphi(\hat{s}^m_i)\}_{i=t}^{t+I}$, reward $\{\hat{r}_{i} \sim p_{\phi}\left(\hat{r}_{i} \mid s^{m}_{i}\right)\}_{i=t}^{t+I}$.
                \STATE  Update world model parameters $\phi$ via Eq. \ref{World model loss}. Update critic parameters $\varphi$ via Eq. \ref{Critic loss} and actor parameters $\theta$ via Eq. \ref{Actor loss}. 
            \ENDFOR
            \LCOMMENT{ Multi-source data collection}
            \FOR{time step $t=1$ \TO $T$}
                \STATE Compute state $s_t$.
                \STATE Extract $s^m_t$ from $s_t$.
                \STATE Take action $a_t \sim \pi_\theta(\hat{a}_t|s^m_t)$.
                \STATE Collect ${r_t,o_{t+1},m_{t+1}}$ from the environment.
            \ENDFOR
            \IF{Collision}
                \STATE Add transition to collision dataset\\ $D_{col} \leftarrow D_{col} \cup \left\{(o_t,m_t,a_t,r_t)\right\}_{t=T-4L}^{T}$.
            \ELSIF{Outlane}
                \STATE Add transition to outlane dataset\\ $D_{out} \leftarrow D_{out} \cup \left\{(o_t,m_t,a_t,r_t)\right\}_{t=T-4L}^{T}$.
            \ENDIF
            \STATE Add transition to common dataset\\ $D_{com} \leftarrow D_{com} \cup \left\{(o_t,m_t,a_t,r_t)\right\}_{t=1}^{T}$.
        \ENDWHILE
    \end{algorithmic}
\end{algorithm}

%===============================================================================

\section{Experiments}
\label{sec:Experiments}

\subsection{Simulation Environment Setup and Details}
To further validate our method's outperformance, we use CARLA~\cite{dosovitskiy2017carla}, a simulator of autonomous driving research, to conduct extensive experiments. CARLA has rich weather conditions and maps that can support us in testing our agents in different weather to verify environmental adaptability. All experiments were conducted on the NVIDIA RTX4090 and Intel i7-13700KF.

\begin{figure}[ht]
    \centering
    \includegraphics[scale=0.3]{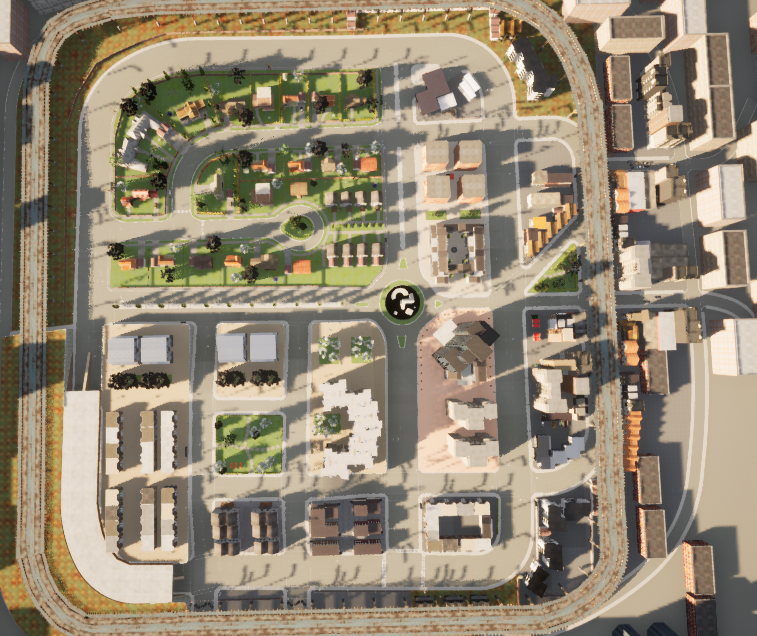}
    \caption{\label{6} Bird-view of town 3 in CARLA.}
    \vspace{-0.01cm}
\end{figure}

The map we use is town 3 which is a complex urban environment shown in Fig. \ref{6}. The map is very close to the real city road environment, with a variety of scenarios such as tunnels, intersections, roundabouts, curves, turnaround bends, etc. The task is to drive along the complex urban environment to get rewards as high as possible in 1000 time steps without going out of bounds and colliding with 100 surroundings. The CARLA operates synchronously at 10Hz meaning a time step of 0.1s. The action is composed of throttle and steering.

In terms of sensors, we use lidar and front view camera to obtain point clouds and camera images as inputs. Lidar is 64 lines with a scanning range of $50m$ and is positioned at a height of $1.8m$. The camera has 110° field of view (FOV) and is positioned at a height of $1.7m$. The size of the camera, lidar, and mask is set to $o_{t},m_{t} \in \left\lbrack {0,~255} \right\rbrack^{128 \times 128 \times 3}$.

We train the world model with batch size $B = 16$ and batch length $L = 16$. For training the policy network, the imagined horizon $I$ is set to 8. Stochastic state $z_{t} \in R^{32 \times 32}$ and deterministic state $h_{t} \in R^{2048}$. Model learning rate is set to $3 \times 10^{- 5}$ and actor-critic learning rate is set to $1 \times 10^{- 5}$. The optimizer we use is AdamW~\cite{loshchilovdecoupled}. The discount factor $\gamma$ is 0.99. The TD-$\lambda$ factor $\lambda$ is 0.95. The entropy factor $\eta$ is $1 \times 10^{- 4}$. SEM2 and baseline DreamerV2 use the exact same parameters to ensure fairness.

\subsection{Reward Function}
Our reward function is similar to Chen et al.~\cite{chen2021interpretable}, which can be represented as:

\begin{equation}
\begin{aligned}
r~ = ~200~r_{col} + v_{lon} + 10~r_{fast} + r_{out}\\
- 5\alpha^{2} + 0.2~r_{lat} + r_{cte} + 10~r_{safe} - 0.1.
\end{aligned}
\end{equation}

The term $r_{col}$ is set to -1 if a collision occurs. Collision, as the most avoidable condition of automatic driving, is assigned a great penalty. Ego vehicles can even take outlane behaviors to avoid collisions when necessary. The term $v_{lon}$ is the longitudinal speed ego vehicle. The term $r_{fast}$ is set to -1 when the speed of the ego vehicle exceeds desired speed ($8m/s$). The term $r_{out}$ is set to -1 when the cross-track error (CTE) exceeds the threshold ($2m$). The punishment should not be excessive, or else the ego vehicle tends to fall into the local optimal solution waiting for the exhaustion of the episode. The $\alpha$ is the steering angle in rad. The term $r_{lat}$ is related to lateral acceleration computed as $-\vert\alpha\vert v_{lon}^{2}$. The term $r_{cte}$ is minus CTE to keep the ego vehicle in the center of the lane. The term $r_{safe}$ is set to -1 when the distance between the ego vehicle center and front vehicle center is less than the threshold ($8m$) and the brake of ego vehicle is less than 0.7. In the last term, we introduce a constant punishment to make the vehicle stay in motion.

\subsection{Network Architecture}
The parametrized networks in SEM2 include the RSSM, the semantic filter $S(s^m_t|h_t,z_t)$, the mask predictor $p_(\hat{m}_t|s^m_t)$, the observation predictor $p\left(\hat{o}_{t} \mid h_{t}, z_{t}\right)$, the reward predictor $p\left(\hat{r}_{t} \mid s^{m}_{t}\right)$, the value network $v(s^m_t)$, and the policy network $\pi(\hat{a}_t|s^m_t)$.

The encoder in RSSM consists of 5 convolutional layers ((32, 4, 2), (64, 4, 2), (128, 4, 2), (256, 4, 2), (256, 4, 2)) each characterized by a tuple means (filters, kernel size, strides). The RSSM uses a GRU with 200 units as a deterministic path and all other functions are implemented as a two-layer MLP of size 1000 with ELU activations. The semantic filter consists of a two-layer MLP of size 1024 with ELU activations and a decoupling header containing an FC head which is implemented as an FC layer and a categorical head which is implemented as an FC layer and a one-hot categorical distribution. The mask and observation predictor consists of an FC and 5 transposed convolutional layers ((1024, 5, 2), (512, 5, 2), (128, 5, 2), (64, 6, 2), (32, 6, 2)). The reward predictor, value network, and policy network all consist of a five-layer MLP of size 400 with ELU activations.

\subsection{Multi-source Data Collection}
\begin{figure}[!t]
\centering
\includegraphics[scale=0.15]{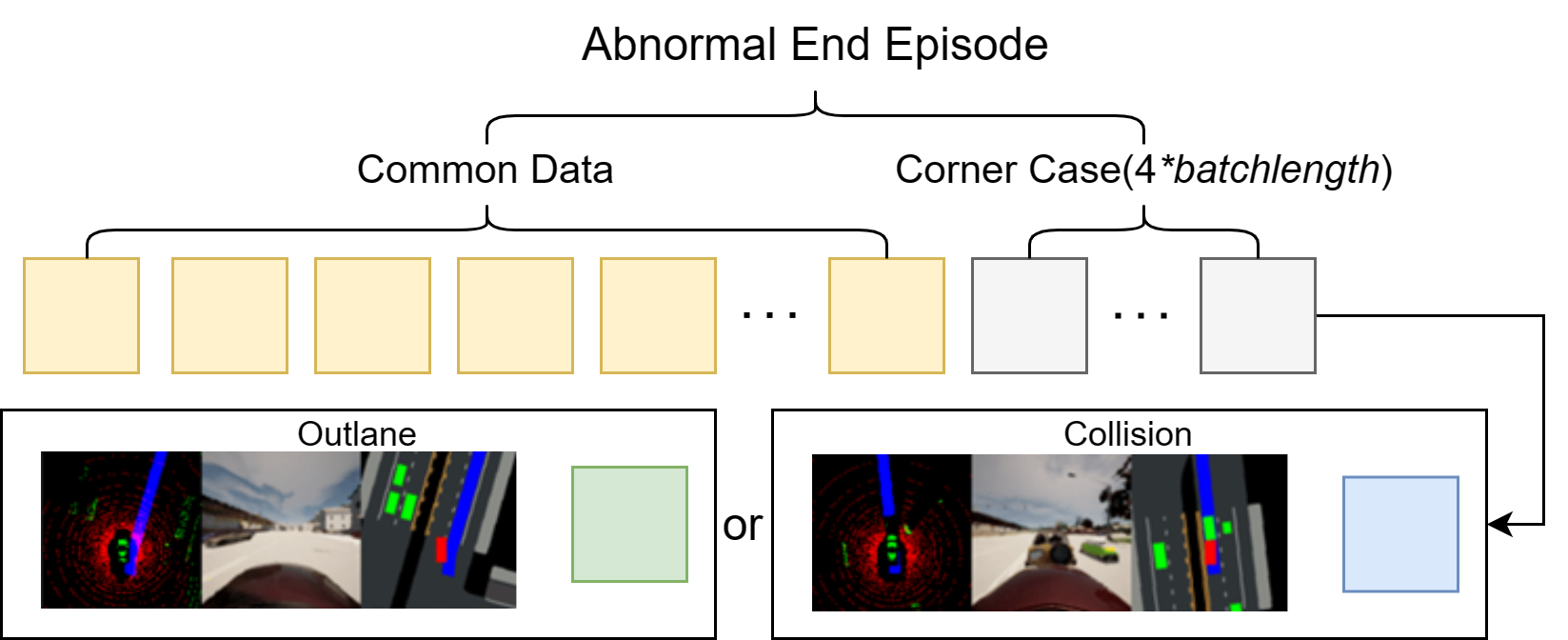}
\caption{\label{7} Multi-source data collection. The data from abnormal end episodes can be categorized into common data and corner cases. The corner cases can be further divided into collision data and outlane data, which are highly relevant to collision and outlane scenarios. The multi-source data sampler divides the data into different replay buffers based on their types.}
\end{figure}

The corner case is defined as an abnormal end case during the data collection process. When an abnormal end occurs, the CTE of the ego vehicle is greater than $2m$ and thus out of bounds or there is a collision with a surrounding object. These two cases shown in Fig. ~\ref{7} correspond to outlane and collision. When an abnormal end occurs, the multi-source data sampler puts the last $4\times L$ steps of the episode into the replay buffer according to the type of end. We evaluate the agent every 20,000 steps to adaptively adjust the data distribution. The  max corner data probability $P_{maxcor}$ is set to $0.5$.

%===============================================================================

\section{Experimental Results}
\label{sec: Experimental Results}
To properly evaluate SEM2 as a reinforcement learning algorithm and an end-to-end autonomous driving system, we considered multiple factors. From a reinforcement learning perspective, we looked at the average return as well as the evaluating curves to assess the sample efficiency of the algorithm. From the perspective of autonomous driving, we tested the driving performance including success rate, collision rate, outlane rate, incomplete rate and average driving distance of each algorithm in both random and selected road sections. In addition, we tested the robustness of the algorithm in different weather with salt-pepper noise to assess the effectiveness of the semantic filter. We also made an ablation study to demonstrate the significance of the semantic filter in performance improvement. And then an ablation experiment is conducted to validate the effectiveness of adaptive multi-source data distribution. Finally, we provide an overview of the various driving scenarios encountered by the autonomous driving system and describe the behavior of the system.

\subsection{Baseline Algorithms and Variants of Proposed Method} 
We compare the RL algorithms baselines and variants of our proposed method:
\begin{enumerate}
\item{\textit{SAC}~\cite{haarnoja2018soft}:} SAC, a model-free maximum entropy deep RL algorithm that learns a stochastic policy and has achieved leading results in many benchmarks. We choose SAC as the model-free baseline to highlight the sample efficiency of the model-based algorithm.
\item{\textit{DreamerV2}~\cite{hafner2021mastering}:} DreamerV2 is a model-based RL algorithm that acquires behaviors through imagined trajectories in the efficient latent space of an advanced RSSM. DreamerV2 as a model-based baseline demonstrates the high performance of SEM2. To ensure fairness, the DreamerV2 in our setting also has a mask predictor which facilitates the mask.
\item{\textit{SEM2 without multi-source}:} To verify the effectiveness of the multi-source data sampler, we set up a variant of SEM2, which is without multi-source data training. This is an ablation version of SEM2.
\item{\textit{SEM2 (Proposed)}:} \textbf{SEM}antic \textbf{M}asked recurrent world model which we proposed, the full version which uses multi-source data training.
\end{enumerate}

\begin{figure}[!t]
    \centering
    \includegraphics[scale=0.11]{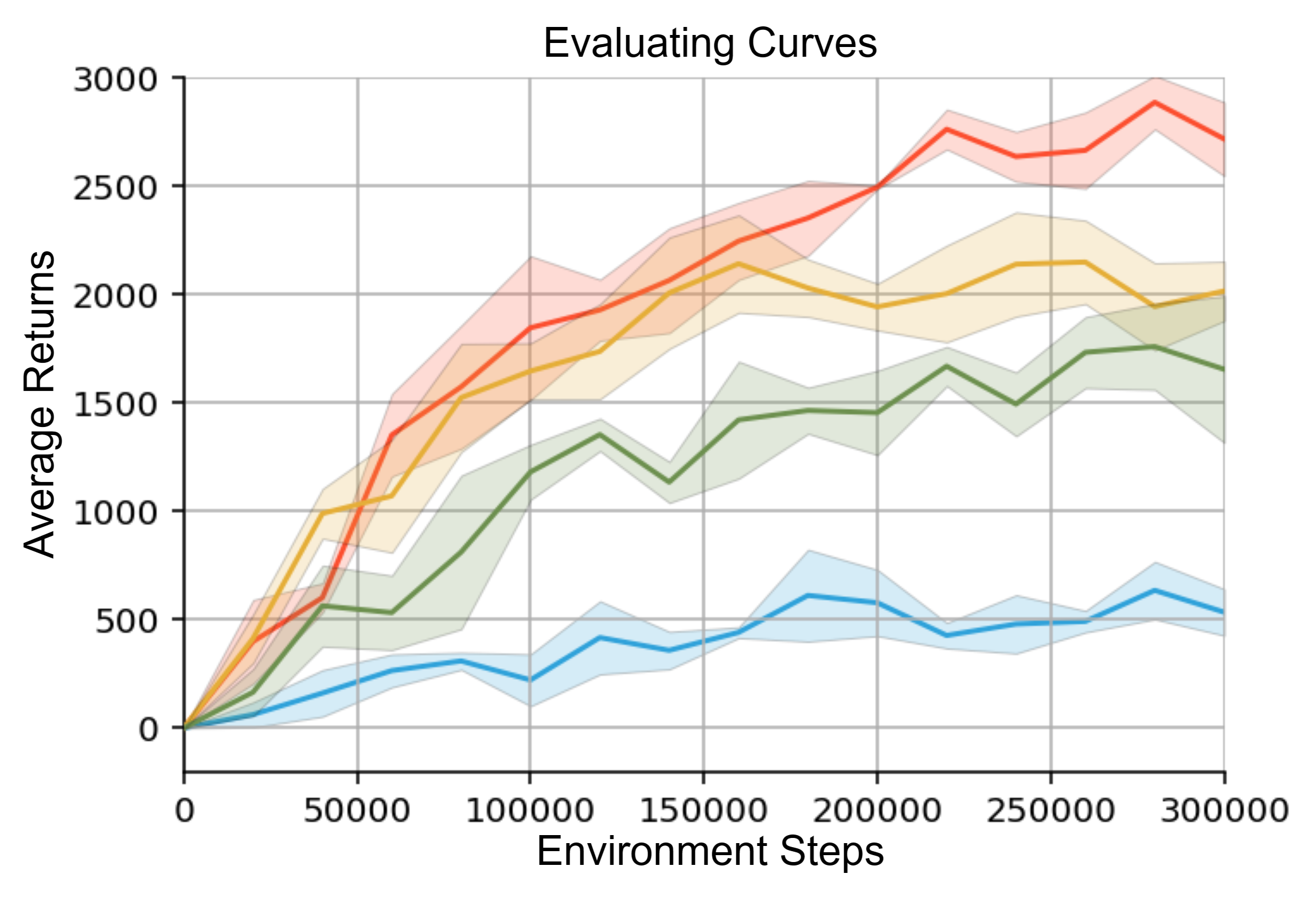}
    \includegraphics[scale=0.08]{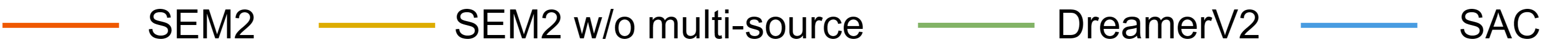}
    \setlength{\abovecaptionskip}{-0.05cm}
    \vspace{-0.03cm}
    \caption{\label{8} Evaluating curves. The evaluating curves record the average returns for taking 30 episodes at checkpoints which are recorded every 20,000 steps. In the evaluation, In each new episode, vehicles are randomly placed on the map. All these average returns are calculated with 3 trials. The shaded area indicates standard deviation.}
\end{figure}

\subsection{Evaluating Curves} 
All experiments' returns are average of 3 trials. Each trial is executed for 300,000 environmental steps. Evaluating curves in Fig. \ref{8} show the performances of SAC, DreamerV2, SEM2 without multi-source training, and SEM2 training in town 3 and clear noon. Our results show that SEM2 has higher sample efficiency and achieves a higher average return than baselines. The curves also clearly demonstrate that multi-source training is beneficial in improving the average return.

\sethlcolor{yellow}
\subsection{Driving performance}

SEM2 is an end-to-end autonomous driving system. In addition to the average return, which serves as an evaluation metric, success rate, collision rate, outlane rate, incomplete rate and average driving distance that effectively reflect driving performance are also employed as evaluation criteria for the algorithms. Consequently, we conducted two sets of tests. In these tests, we used the checkpoints with the highest average returns and tested 100 episodes to record metrics.

\begin{figure}[ht]
    \centering
    \includegraphics[scale=0.32]{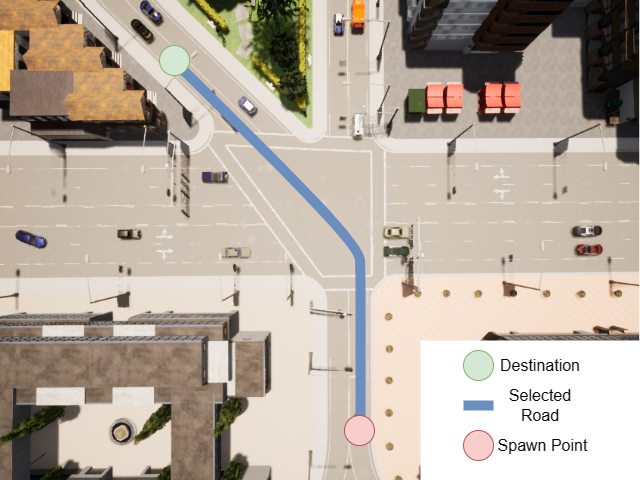}
    \caption{\label{selected intersection} Bird-view of the selected intersection.}
    \vspace{-0.01cm}
\end{figure}

\begin{enumerate}
\item{\textit{Random road sections}: Following a setting similar to the training phase, the ego vehicle is initiated at random positions within town 3 and travels along random roads for 500 time steps. If the vehicle gets a return greater than 300 at the end of 500 steps it is recorded as successful, otherwise it is recorded as incomplete.}
\item{\textit{Selected road sections}: In this setting, the ego vehicle is initiated at selected positions and travels along selected roads to reach the destination. The SAC is not tested in this setting due to its poor performance only being able to travel along a straight line. We tested the performance of the agents to pass through the selected intersection which is shown in Fig. \ref{selected intersection}.}
\end{enumerate}

\begin{table*}[htbp]
\vspace{-10pt}
\renewcommand\arraystretch{1.5}
\centering
\caption{Driving performance in random road sections}
\centerline{
\begin{tabular}{l|ccccc}
    \toprule
    Algorithm & Success rate & Collision rate & Outlane rate & Incomplete rate& Average driving distance \\
    \midrule
    SAC & $18.0\pm1.0\%$ & $8.0\pm1.0\%$ & $33.0\pm3.0\%$ & $41.0\pm3.0\%$ & $55.5\pm4.2m$\\
    DreamerV2 & $51.0\pm2.9\%$ & $29.0\pm5.9\%$ & $19.3\pm3.9\%$ & $0.7\pm0.5\%$ & $209.1\pm10.5m$ \\
    SEM2 w/o multi-source & $64.7\pm2.9\%$ & $26.3\pm1.3\%$ & $8.0\pm2.9\%$ & $1.0\pm1.0\%$ & $234.4\pm19.0m$\\
    \cellcolor{codegray}SEM2 & \cellcolor{codegray} ${\bf79.7\pm2.6\%}$ & \cellcolor{codegray}  ${\bf14.3\pm3.3\%}$ & \cellcolor{codegray} ${\bf5.3\pm0.5\%}$ & \cellcolor{codegray} ${\bf0.3\pm0.5\%}$ & \cellcolor{codegray} ${\bf243.5\pm3.2m}$ \\
    \bottomrule
\end{tabular}
}
\label{driving performance in random road sections}
\vspace{-0.05cm}
\end{table*}

\begin{table*}[htbp]
\renewcommand\arraystretch{1.5}
\centering
\caption{Driving performance in the selected intersection}
\centerline{
\begin{tabular}{l|ccccc}
    \toprule
    Algorithm & Success rate & Collision rate & Outlane rate & Incomplete rate& Average driving distance \\
    \midrule
    DreamerV2 & $71.0\pm5.7\%$ & $29.0\pm5.7\%$ & $0.0\pm0.0\%$ & $0.0\pm0.0\%$ & $55.5\pm2.7m$ \\
    SEM2 w/o multi-source & $90.3\pm4.1\%$ & $9.7\pm4.1\%$ & $0.0\pm0.0\%$ & $0.0\pm0.0\%$ & $67.8\pm1.6m$\\
    \cellcolor{codegray}SEM2 & \cellcolor{codegray} ${\bf96.7\pm1.7\%}$ & \cellcolor{codegray}  ${\bf3.3\pm1.7\%}$ & \cellcolor{codegray} ${\bf0.0\pm0.0\%}$ & \cellcolor{codegray} ${\bf0.0\pm0.0\%}$ & \cellcolor{codegray} ${\bf70.8\pm0.3m}$ \\
    \bottomrule
\end{tabular}
}
\label{driving performance in random road sections}
\vspace{-0.05cm}
\end{table*}

Both sets of tests confirmed the outstanding driving performance of SEM2 and demonstrated the contributions of the semantic mask filter and the multi-source data sampler.

\subsection{Effect of Semantic Filter}
\begin{figure*}[htbp]
    \centering
    \subfloat[Clear noon]{
    \centering
    \includegraphics[scale=0.08]{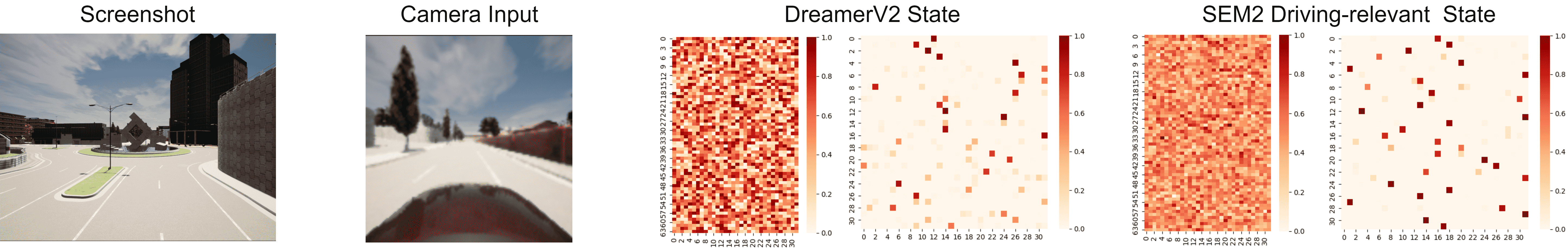}
    }

    \vspace{-0.42cm}
    
    \subfloat[Hard rain noon]{
    \centering
    \includegraphics[scale=0.08]{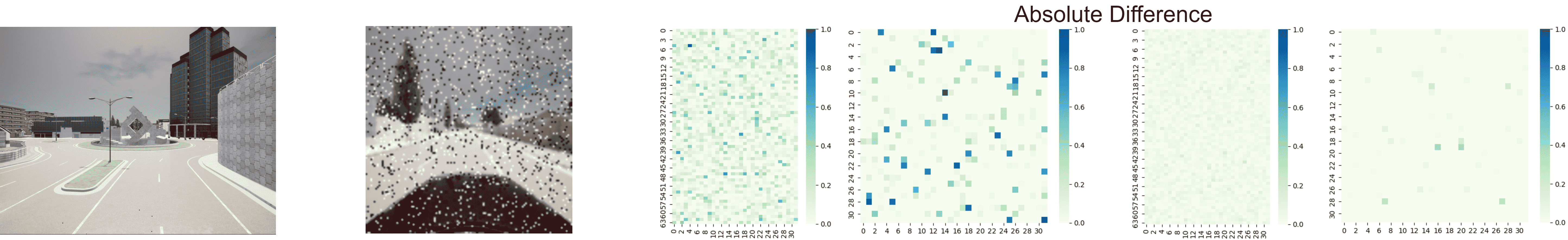}
    }

    \vspace{-0.42cm}
    
    \subfloat[Wet sunset]{
    \centering
    \includegraphics[scale=0.08]{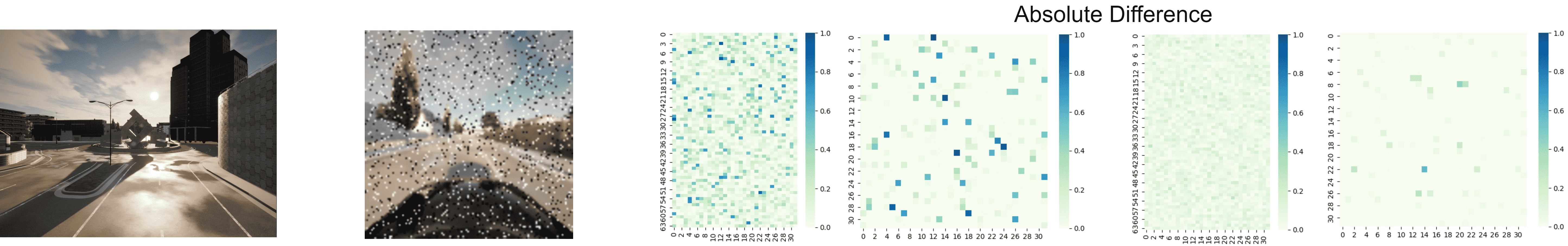}
    }
    
    \centering
    \caption{\label{weather} We conducted a series of tests under varying weather conditions, incorporating salt-pepper noise as input permutation. The first column is screenshots of the CARLA simulation, while the second column displays the camera input. The third and fourth columns in first raw depict heatmaps of the state $\{h_{t},z_{t}\}$ within the features of DreamerV2 and the state $\{h^{m}_{t},z^{m}_{t}\}$ within the driving-relevant features of SEM2 in clear noon, respectively. The absolute difference between the heatmaps under the conditions of hard rain noon and wet sunset with salt-pepper noise condition, as well as the heatmap under clear noon, is illustrated in the third and fourth columns of the second and third rows. The deterministic state is a one-dimensional vector of size 2048, for ease of visualization, we have normalized it to [0,1] and represented it as a 64×32 heatmap. The probability of stochastic state is represented as a 32×32 heatmap.}
    \vspace{-10pt}
\end{figure*}

Irrelevant information in the image input can impact autonomous driving performance. The semantic filter in SEM2 extracts driving-relevant features that contain less irrelevant information, allowing SEM2 to perform well in various weather conditions and improving its robustness of input perturbations. To verify this, we evaluated our agents in two new weather conditions they had not encountered before: wet sunset and hard rain noon. Moreover, as the impact of weather is relatively less challenging for an algorithm that exhibits robustness, we incorporated salt-pepper noise into the camera input (shown in Fig. \ref{weather}). This noise is designed to make the camera input resemble images captured under extreme rainstorm conditions. The probability of salt-pepper noise is $0.1$. Tab. \ref{different weather} shows that SEM2 outperforms the baselines in all these new weather conditions with the salt-pepper noise. The numbers in parentheses in table indicate the change percentage of the average returns in the robustness test compared to the training set. The results of SEM2 and SEM2 without multi-source training also indicate that the improvement in robustness is mainly due to the use of the semantic mask, while the multi-source training method has negligible influence on robustness. 

Fig. \ref{weather} also presents states' heatmaps illustrating the absolute difference between noon and bad weather. 
The red heatmap represents states under noon conditions. The blue heatmap represents the absolute difference between states under varying weather conditions and noon. It is evident that SEM2's driving-relevant states are less affected by input perturbations, whereas DreamerV2's states undergo significant changes, indicating the presence of a considerable amount of irrelevant information associated with visual inputs.

To confirm that using driving-relevant features which not contain the weather information still enables the agent to drive robustly when the environment changes impact driving, we conducted an additional experiment. We evaluated various algorithms under hard rain conditions with degraded braking efficiency. Tab. \ref{different braking efficiency} revealed that both SEM2 and DreamerV2 demonstrate robustness against the reduction in brake efficiency during rainy conditions. However, in the case of excessive attenuation of brake efficiency, the ego vehicle cannot stop in time when the vehicle in front brakes suddenly, resulting in performance degradation. SEM2 without the multi-source algorithm exhibits a greater performance decline due to poorer control of following distance when braking efficiency decreases, but it still outperforms DreamerV2. 
The superior performance of SEM2 w/o multi-source over DreamerV2 can be attributed to its higher average return in the standard setup. So semantic filter does not inherently enhance the agent's robustness to changes in environmental dynamics. The results of SEM2 demonstrate training with multi-source data allows the agent to handle corner cases better and maintain an effective safety distance, even as brake efficiency decreases.

\begin{table*}[htbp]
\renewcommand\arraystretch{1.5}
\centering
\caption{The average returns under different weather conditions with salt-pepper noise \\(clear noon is the training environment without salt-pepper noise)}
\centerline{
\begin{tabular}{l|c|cc}
    \toprule
    Algorithm & Clear Noon & Hard Rain Noon  & Wet Sunset \\
    \midrule
    SAC & $534.9\pm122.3$ & $436.8\pm95.1(-18.3\%)$ & $461.1\pm105.5(-13.8\%)$ \\
    DreamerV2 & $1655.9\pm314.8$ & $1435.1\pm255.0(-13.3\%)$ & $1390.4\pm338.6(-16.0\%)$ \\
    SEM2 w/o multi-source & $2017.1\pm127.1$ & $2081.1\pm176.4(+3.9\%)$ &  $2098.2\pm163.5(+4.9\%)$\\
    \cellcolor{codegray}SEM2 & \cellcolor{codegray} ${\bf2719.1\pm158.4}$ & \cellcolor{codegray}  ${\bf2757.6\pm156.0(+1.4\%)}$ & \cellcolor{codegray} ${\bf2775.9\pm120.2(+2.1\%)}$ \\
    \bottomrule
\end{tabular}
}
\label{different weather}
\vspace{-0.05cm}
\end{table*}

To demonstrate the significance of the semantic filter in performance improvement, an additional ablation study is conducted to isolate the effect of the semantic filter and the mask predictor. Fig. ~\ref{ablation} presents the results of the ablation study. It is observed that the presence of the mask predictor does not have a significant impact on the performance improvement, as evidenced by the evaluation curves of DreamerV2 with mask predictor and DreamerV2 without mask predictor. This suggests that the mask predictor primarily contributes to interpretability, rather than performance enhancement. In contrast, the results of the SEM2 model without multi-source training revealed the critical role played by the semantic filter in leveraging high-dimensional semantic information.

We also conducted experiments on the representation forms of stochastic states. In the labels of Fig.~\ref{ablation}, Gaussian denotes continuous representations generated by 32 Gaussian distributions, while One-hot refers to discrete representations generated by 32 one-hot categorical distributions. The results demonstrate that employing one-hot representation facilitates optimization, contributing to improved sample efficiency.

\subsection{Distribution of Multi-source Data}
\begin{figure*}[htbp]
    \begin{minipage}[t]{0.49\textwidth}
    \centering
    \includegraphics[scale=0.11]{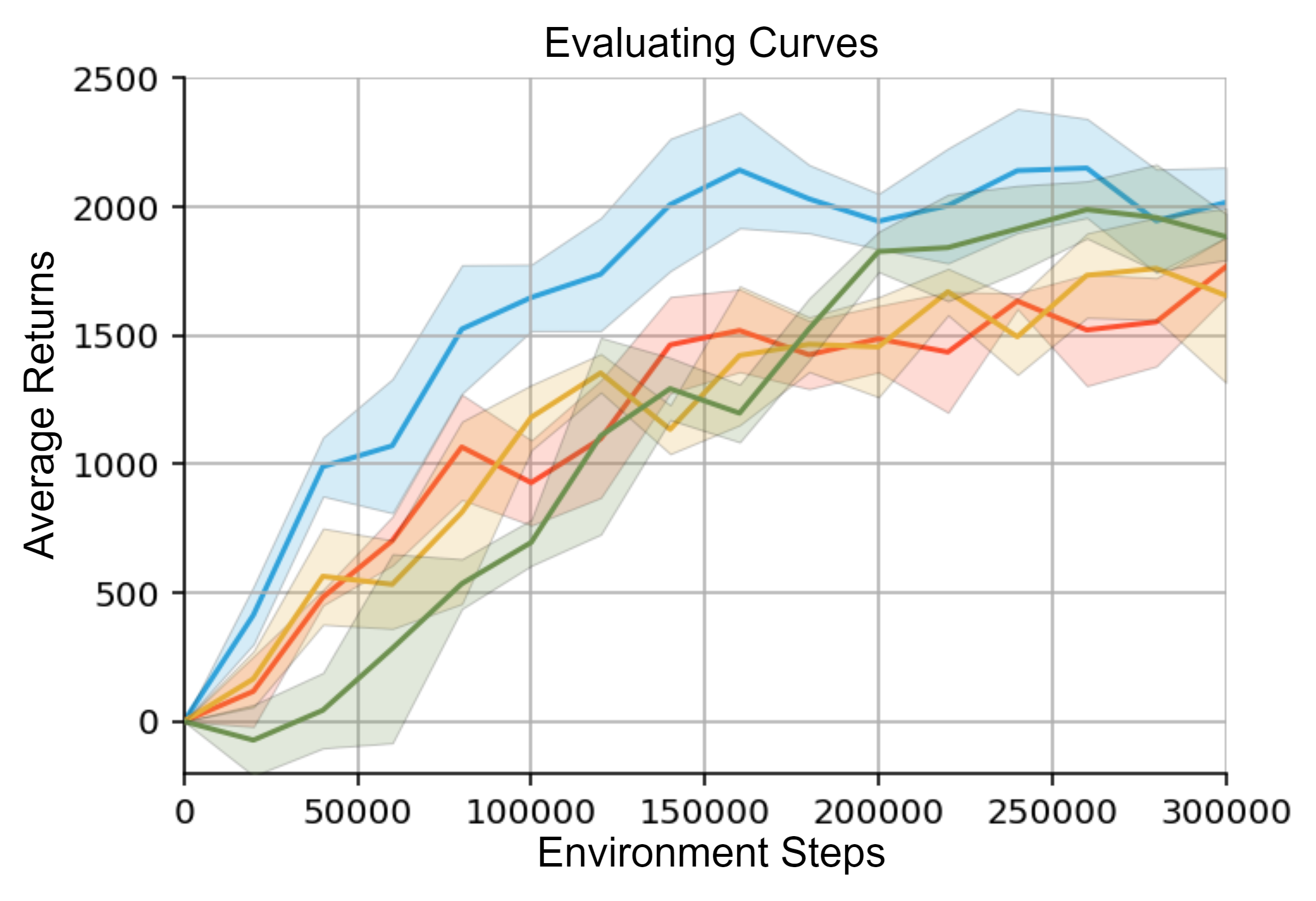}
    \includegraphics[scale=0.065]{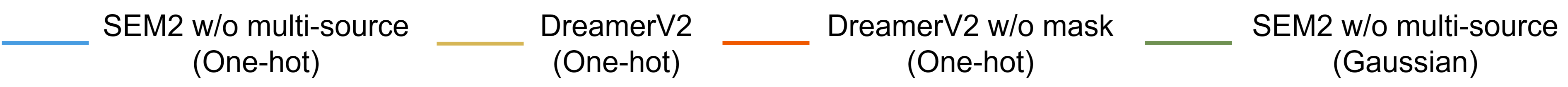}
    \setlength{\abovecaptionskip}{-0.05cm}
    \caption{\label{ablation} Evaluating curves of the effect of semantic filter and representation form ablation study. The curves record the average returns for taking 30 episodes. The vehicles are randomly relocated on the map and the results are averaged over 3 trials. The shaded area indicates standard deviation.}
    \end{minipage}
    \hfill
    \begin{minipage}[t]{0.49\textwidth}
    \centering
    \includegraphics[scale=0.11]{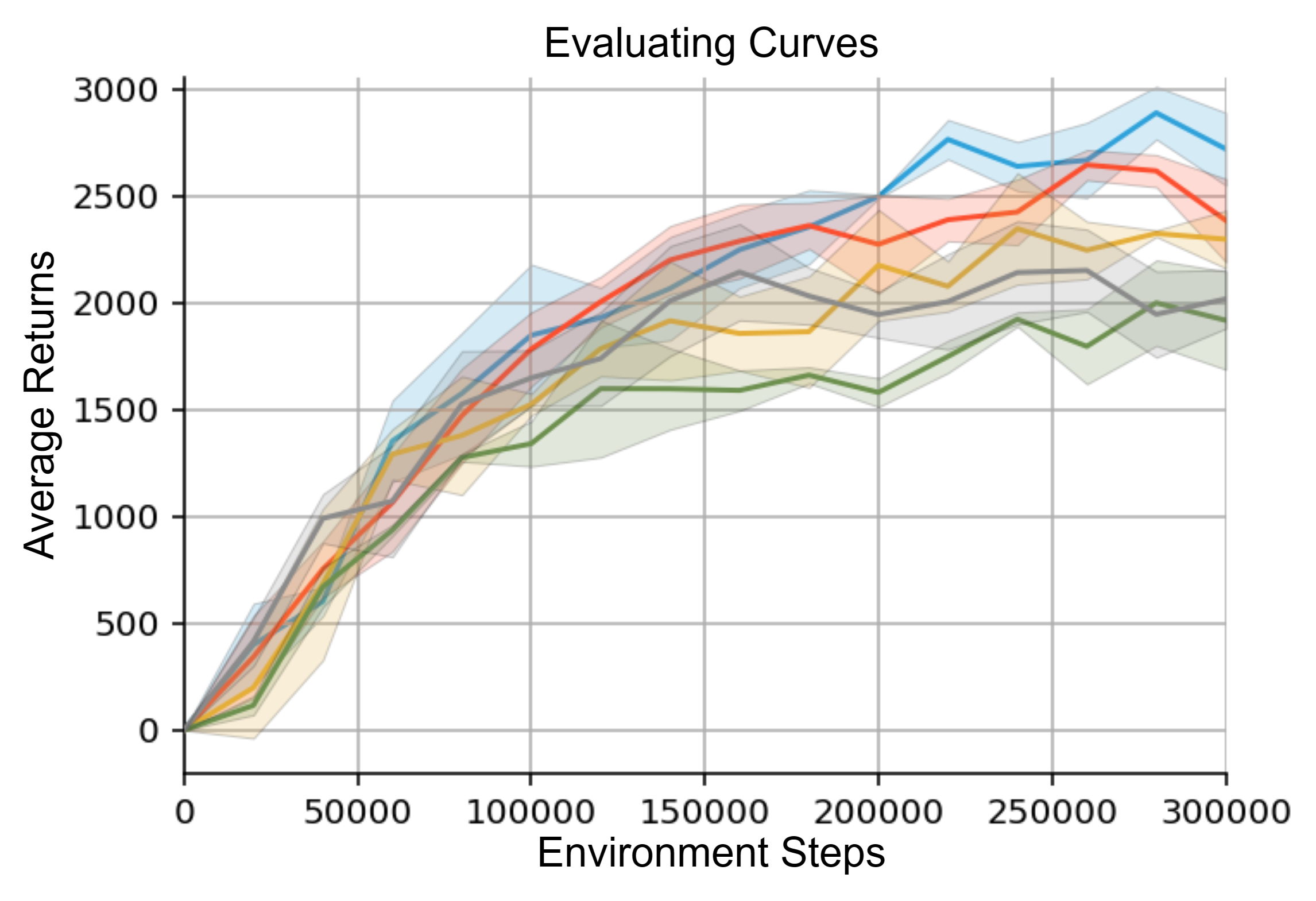}
    \includegraphics[scale=0.08]{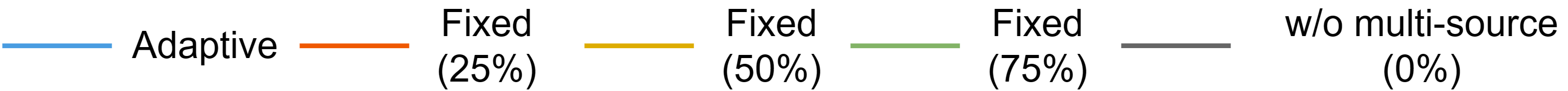}
    \setlength{\abovecaptionskip}{-0.05cm}
    \vspace{-0.03cm}
    \caption{\label{multi-source data} Evaluating curves of different multi-source data distributions. The curves record the average returns for taking 30 episodes. The vehicles are randomly relocated on the map and the results are averaged over 3 trials. The shaded area indicates standard deviation.}
    \end{minipage}
\end{figure*}

To validate the effectiveness of adaptive multi-source data distribution, we conducted an ablation experiment. This experiment compares the average returns of the adaptive multi-source data distribution, the fixed corner data distribution (25\%,50\%,75\%) and without the multi-source data training (0\%). The results of the experiment are shown in Fig. \ref{multi-source data}.

The findings from this ablation study reveal a positive influence on performance enhancement across all adaptive data distribution, 25\% and 50\% fixed corner data distribution. Notably, employing 75\% corner data engenders an imbalance of data by an insufficient of normal data, thereby precipitating a decrement in the average return.

\begin{table*}[htbp]
\renewcommand\arraystretch{1.5}
\centering
\caption{The average returns under hard rain noon with salt-pepper noise and different braking efficiency}
\centerline{
\begin{tabular}{l|c|ccc}
    \toprule
    Algorithm & 100\% & 90\% & 80\% & 70\% \\
    \midrule
    SAC & $436.5\pm95.1$ & $473.6\pm96.3(+8.0\%)$ & $482.2\pm90.7(+9.9\%)$ & $501.1\pm91.9(+14.0\%)$\\
    DreamerV2 & $1435.1\pm255.0$ & $1444.9\pm299.9(+0.1\%)$ & $1386.5\pm256.0(-3.3\%)$ & $1384.5\pm264.1(-3.6\%)$\\
    SEM2 w/o multi-source & $2081.1\pm176.4$ &  $2046.4\pm221.6(-1.6\%)$ & $1858.2\pm233.9(-10.7\%)$ & $1790.5\pm139.3(-14.0\%)$\\
    \cellcolor{codegray}SEM2 & \cellcolor{codegray} ${\bf2757.6\pm156.0}$ & \cellcolor{codegray}  ${\bf2729.8\pm179.1(-1.0\%)}$ & \cellcolor{codegray} ${\bf2681.4\pm118.9(-2.7\%)}$ & \cellcolor{codegray}${\bf2700.9\pm171.7(-2.1\%)}$\\
    \bottomrule
\end{tabular}
}
\label{different braking efficiency}
\vspace{-0.05cm}
\end{table*}

\begin{figure*}[!b]
    \centering
    \subfloat[Straight]{
    \centering
    \includegraphics[scale=0.19]{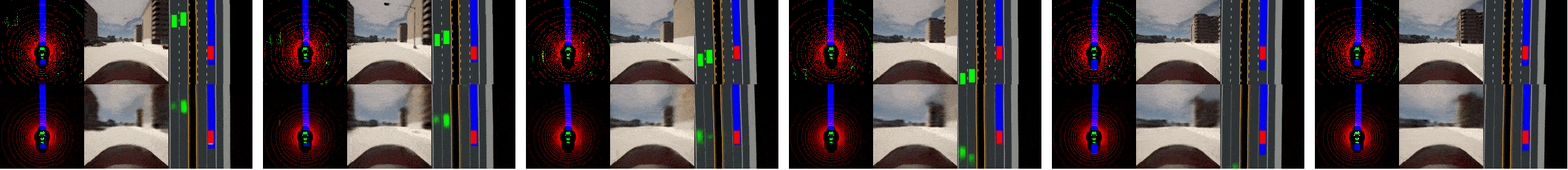}
    }
    
    \subfloat[Curve]{
    \centering
    \includegraphics[scale=0.19]{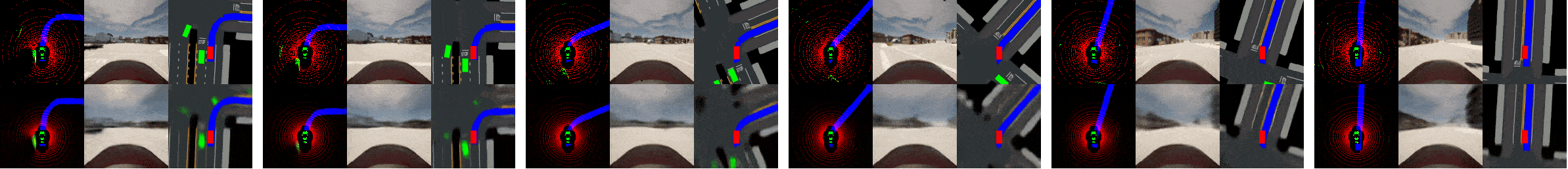}
    }
    
    \subfloat[Vehicle following]{
    \centering
    \includegraphics[scale=0.19]{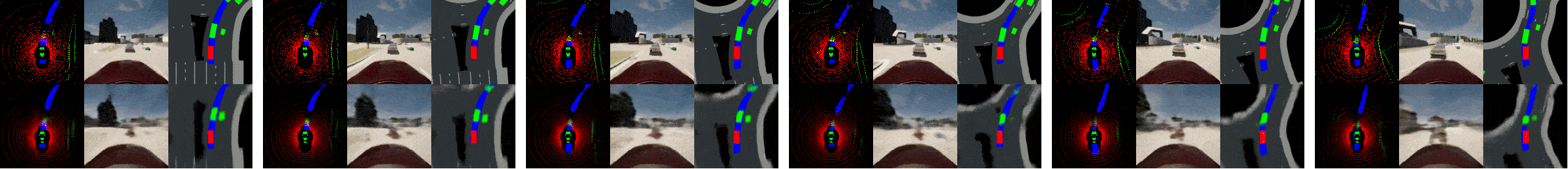}
    }
    
    \subfloat[Intersection]{
    \centering
    \includegraphics[scale=0.19]{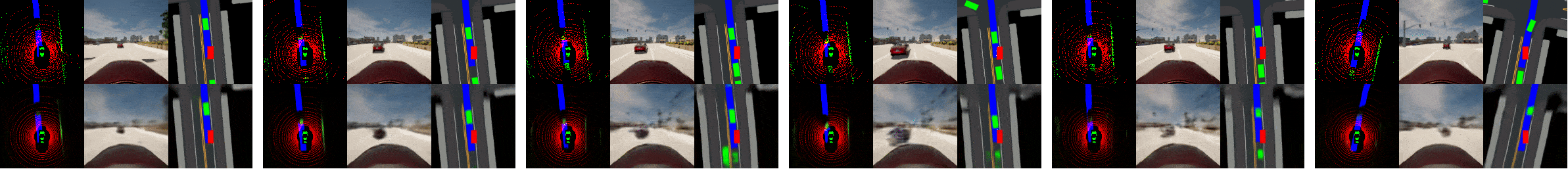}
    }
    
    \centering
    \caption{\label{10} Cases in complex town environments. The figure presented the processing of SEM2 across four distinct road scenarios. Each image, from left to right, displays the lidar input, camera input, and semantic mask. The top row shows the ground truth, while the bottom row shows the reconstructed image.}
\end{figure*}

\subsection{Cases Interpretation}
Once trained, SEM2 is well equipped to handle many cases in town environments shown in Fig. \ref{10}. We analyzed the agent's behavior in these cases to understand its capabilities.

\begin{enumerate}
\item{\textit{Straight}:} The most common driving situation is traveling straight, and the agent can handle it effectively even in the early stages of training. In this case, the autonomous system simply maintains a high-speed cruise and stays within its lane.
\item{\textit{Curve}:} In town roads, there are various types of curves. To ensure that the vehicle stays on the centerline of the curve, the agent adjusts the speed and steering appropriately when entering the curve. 
\item{\textit{Vehicle following}:} The urban environment has a dense traffic flow, and following the vehicle is a common working condition. When there is a vehicle in front of the ego vehicle, the agent will automatically follow the vehicle in front and perform acceleration and deceleration to avoid collision with the front vehicle.
\item{\textit{Intersection}:} There are a large number of intersections in the urban environment, and when at a red light, vehicles are waiting in place. The ego vehicle should also slow down and wait in place until other vehicles have passed.
\end{enumerate}

%===============================================================================

\section{Conclusion}
\label{sec:conclusion}
This paper enhances the sample efficiency and robustness of urban end-to-end autonomous driving by proposing a \textbf{SEM}antic \textbf{M}asked recurrent world model (\textbf{SEM2}), which learns the transition dynamics of driving-relevant states with a semantic filter. 
The driving policy is learned by propagating analytical gradients of imagination through the latent dynamics using SEM2 in the condensed latent space. To contribute diverse scene data and prevent modal collapse in corner cases, we propose a multi-source sampler to balance the data distribution that aggregates both common driving situations and multiple corner cases in urban scenes. We trained our framework in the CARLA simulator and compared its performance with state-of-the-art comparison baselines. Experimental results demonstrate that our framework exceeds previous works in terms of sample efficiency and robustness to input permutation.

% \newpage

% $$s$$

% \newpage
%===============================================================================
\bibliographystyle{IEEEtran}
\bibliography{SEM2}{}

\begin{IEEEbiography}
[{\includegraphics[width=1in,height=1.25in,clip,keepaspectratio]{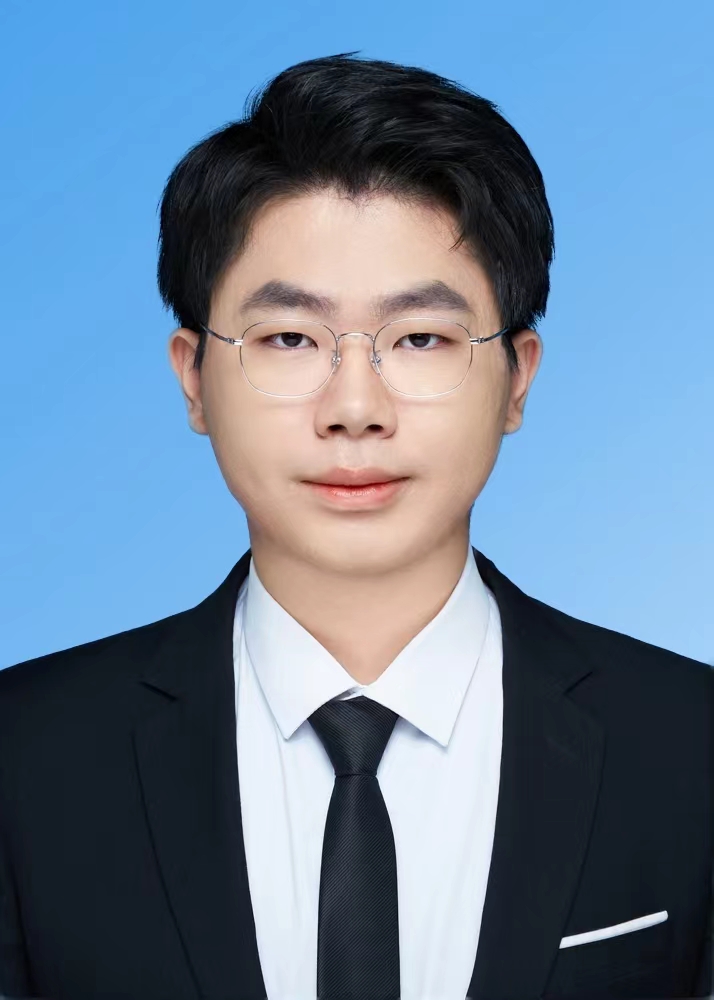}}]
{Zeyu Gao} received the B.S. degree in Intelligent Vehicle Engineering with School of Automotive Engineering, Harbin Institute of Technology, China in 2023. He is currently working toward the M.S. degree in pattern recognition and intelligent system at Institute of Automation, Chinese Academy of Science, China. His research interests include autonomous driving, reinforcement learning, and brain-inspired intelligence.
\end{IEEEbiography}

\begin{IEEEbiography}[{\includegraphics[width=1in,height=1.25in,clip,keepaspectratio]{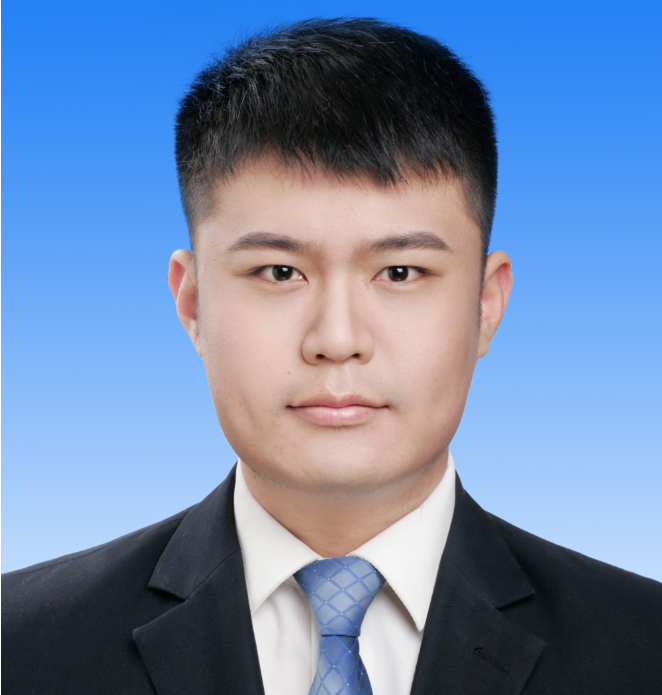}}]
{Yao Mu} obtained the M.Phil Degree from the School of Vehicle and Mobility, Tsinghua University, Beijing. He is pursuing the Ph.D. degree with the department of Computer Science, the University of Hong Kong, Hong Kong. He studied as a visiting student researcher in of Information Technology and Electrical Engineering, ETH Zürich, Zürich, 
Switzerland,
in 2023.  His research interests include embodied AI, reinforcement learning, robotic control, autonomous driving, and computer vision.
\end{IEEEbiography}

\begin{IEEEbiography}[{\includegraphics[width=1in,height=1.25in,clip,keepaspectratio]{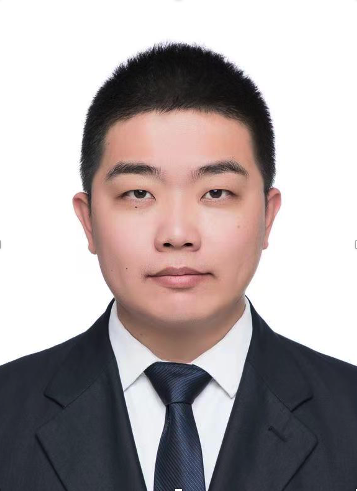}}]
{Chen Chen} received the B.S., M.S., and Ph.D degree in Transportation Engineering in Transportation Engineering from College of Metropolitan Transportation, Beijing University of Technology, Beijing, China, in 2013, 2016 and 2019, respectively. He is currently working as a Post-Doc in the School of Vehicle and Mobility, Tsinghua University, Beijing, China. His research interests include driver behavior analysis, driving safety evaluation, trajectory prediction.
\end{IEEEbiography}

\begin{IEEEbiography}[{\includegraphics[width=1in,height=1.25in,clip,keepaspectratio]{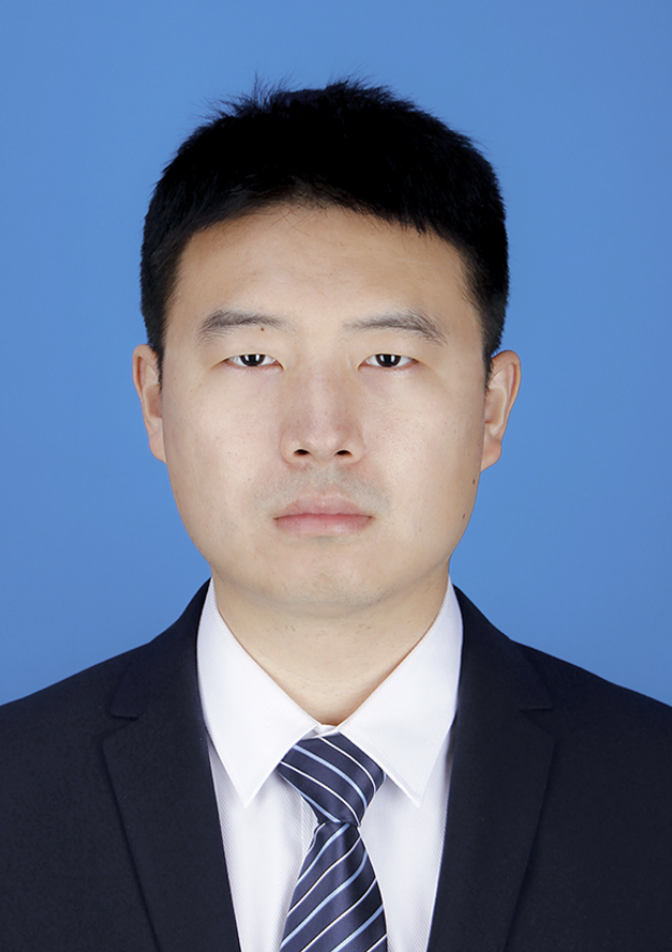}}]
{Jingliang Duan} received the B.S. degree from the College of Automotive Engineering, Jilin University, Changchun, China, in 2015. He studied as a visiting student researcher in Department of Mechanical Engineering, University of California, Berkeley, USA,
in 2019. He received his Ph.D. degree in the School of Vehicle and Mobility, Tsinghua University, Beijing, China, in 2021. He is currently an Associate Professor at the School of Mechanical Engineering, University of Science and Technology Beijing. His research interests include decision and control of autonomous vehicle, reinforcement learning and adaptive dynamic programming, and driver behavior analysis.
\end{IEEEbiography}

\begin{IEEEbiography}[{\includegraphics[width=1in,height=1.25in,clip,keepaspectratio]{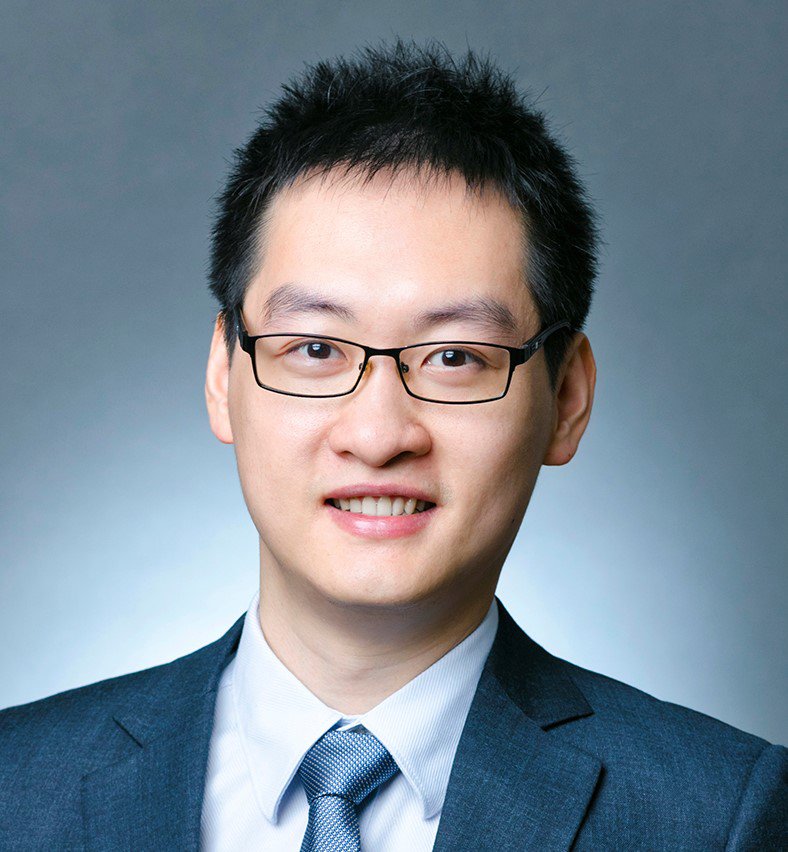}}]
{Ping Luo} (Member, IEEE) received the Ph.D. degree in information engineering from The Chinese University of Hong Kong.
% , in 2014, under the supervision of Prof. Xiaoou Tang and Prof. Xiaogang Wang. 
He was a Postdoctoral Fellow at CUHK from 2014 to 2016. He joined SenseTime Research as a Principal Research Scientist from 2017 to 2018. 
He is currently an Assistant Professor with the Department of Computer Science, The University of Hong Kong. 
He has published more than 100 peer-reviewed papers in top-tier conferences and journals.
% , such as IEEE Transactions on Pattern Analysis and Machine Intelligence (TPAMI), IJCV, ICML, ICLR, CVPR, and NIPS. 
% His work has high impact with 13000 citations according to Google Scholar. 
His research interests include machine learning and computer vision. 
% He has won a number of competitions and awards, such as the First Runner Up at 2014 ImageNet ILSVRC Challenge, the First Place at 2017 DAVIS Challenge on Video Object Segmentation, the Gold medal at 2017 Youtube 8M Video Classification Challenge, the First Place at 2018 Drivable Area Segmentation Challenge for Autonomous Driving, the 2011 HK PhD Fellow Award, and the 2013 Microsoft Research Fellow Award (ten Ph.D. in Asia). He is named one of the young innovators under 35 by MIT Technology Review (TR35) Asia Pacific.
\end{IEEEbiography}

\begin{IEEEbiography}[{\includegraphics[width=1in,height=1.25in,clip,keepaspectratio]{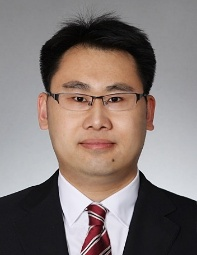}}]
{Yanfeng Lu} (Member, IEEE) received his B.S. degree in Automation from the Harbin Institute of Technology, China in 2010, and his Ph.D. degree from Korea University, Republic of Korea in 2015. He is currently an Associate Professor with the State Key Laboratory of Multimodal Artificial Intelligence Systems, Institute of Automation, Chinese Academy of Sciences, China. His research interests include computer vision, robot perception, and machine learning. 
\end{IEEEbiography}

\begin{IEEEbiography}[{\includegraphics[width=1in,height=1.25in,clip,keepaspectratio]{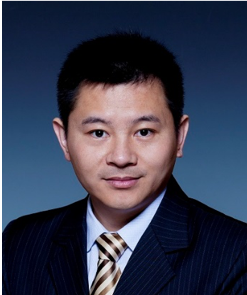}}]
{Shengbo Eben Li} (Senior Member, IEEE) received the M.S. and Ph.D. degrees from Tsinghua University in 2006 and 2009. He worked at Stanford University, University of Michigan, and University of California, Berkeley. He is currently a tenured professor at Tsinghua University. His research interests include intelligent vehicles, driver assistance, reinforcement learning and  optimal control, etc. He is the author of over 100 journal/conference papers, and the co-inventor of over 20 Chinese patents. He is now the IEEE senior member and serves as associated editor of IEEE ITSM and IEEE Trans. ITS, etc.
\end{IEEEbiography}
\end{document}